\documentclass[preprint]{elsarticle}

\usepackage{placeins}
\usepackage{amsmath, amssymb}
\usepackage{color}
\usepackage{lineno}
\usepackage{makecell}
\usepackage{hyperref}       
\usepackage{amsfonts}       
\usepackage{nicefrac}       
\usepackage{microtype}      
\usepackage{doi}
\usepackage{dcolumn}
\usepackage{bm}
\usepackage{caption}
\usepackage{subcaption}
\usepackage{float}
\usepackage{array}
\usepackage{pythonhighlight}
\usepackage{fancyvrb}
\usepackage{xurl}
\usepackage[frozencache=true,cachedir=minted-cache]{minted}
\usepackage{listings}
\usepackage{xcolor}
\usepackage{soul}

\usepackage{algpseudocode}

\newcolumntype{P}[1]{>{\centering\arraybackslash}p{#1}}

\journal{Springer}

\begin{document}

\begin{frontmatter}


\title{Differentiable Neural-Integrated Meshfree Method for Forward and Inverse Modeling of Finite Strain Hyperelasticity}

\author[UMN]{Honghui Du}
\author[UMN]{Binyao Guo}
\author[UMN]{QiZhi He\corref{mycorrespondingauthor}}
\cortext[mycorrespondingauthor]{Corresponding author}
\ead{qzhe@umn.edu}
\address[UMN]{Department of Civil, Environmental, and Geo- Engineering, University of Minnesota, 500 Pillsbury Drive S.E., Minneapolis, MN 55455}

\begin{abstract}


The present study aims to extend the novel physics-informed machine learning approach, specifically 
the neural-integrated meshfree (NIM) method, to model finite-strain problems characterized by nonlinear elasticity and large deformations.
To this end, the hyperelastic material models are integrated into the loss function of the NIM method by employing a consistent local variational formulation. Thanks to the inherent differentiable programming capabilities, NIM can circumvent the need for derivation of Newton-Raphson linearization of the variational form and the resulting tangent stiffness matrix, typically required in traditional numerical methods. 
Additionally, NIM utilizes a hybrid neural-numerical approximation encoded with partition-of-unity basis functions, coined NeuroPU, to effectively represent the displacement solution and streamline the model training process. NeuroPU can also be  used for approximating the unknown material fields, enabling NIM a unified framework for both forward and inverse modeling.
For the imposition of displacement boundary conditions, 
this study introduces a new approach based on singular kernel functions into the NeuroPU approximation, leveraging its unique feature that allows for customized basis functions.
Numerical experiments demonstrate the NIM method's capability in forward hyperelasticity modeling, achieving desirable accuracy, with errors among $10^{-3} \sim 10^{-5}$ in the relative $L_2$ norm,
comparable to the well-established finite element solvers. 
Furthermore, NIM is applied to address the complex task of identifying heterogeneous mechanical properties of hyperelastic materials from strain data, validating its effectiveness in the inverse modeling of nonlinear materials.
To leverage GPU acceleration, NIM is fully implemented on the JAX deep learning framework in this study, utilizing the accelerator-oriented array computation capabilities offered by JAX.

\end{abstract}

\begin{keyword}
Physics-informed machine learning \sep Differentiable programming\sep  Hyperelasticity \sep Hybrid approximation \sep  Inverse modeling  \sep JAX
\end{keyword}

\end{frontmatter}

\section{Introduction}

Modeling hyperelasticity with nonlinear elastic deformations is essential for understanding solid behaviors and predicting  
the mechanical responses of various natural and engineered materials, such as rubbers \cite{beatty1987topics}, 
soft biological tissues \cite{chagnon2015hyperelastic}, and metamaterials \cite{babaee20133d}.
In computational mechanics, finite element (FE) formulations based on finite strain theory have been established to account for geometric and material non-linearities  \cite{holzapfel2002nonlinear,belytschko2014nonlinear}.
However, it is well known that standard finite element schemes remain ineffective in handling mesh entanglement-related issues arising from large material deformations, and thus special numerical remedies, such as adaptive mesh refinement, may be required for simulations.
Given these challenges with FE methods, meshfree (or meshless) methods have gained increasing attention. These methods avoid mesh distortion issues by using node-based discretization to construct locally compact approximation functions.
Typical meshfree methods include the diffuse element method \cite{nayroles1992generalizing}, element-free Galerkin method (EFGM) \cite{belytschko1994element}, reproducing kernel particle method (RKPM) \cite{liu1995reproducing,chen1996reproducing}, HP-clouds method \cite{duarte1996hp}, partition of unity method (PUM) \cite{melenk1996partition}, and meshless local Petrov–Galerkin method (MLPG) \cite{atluri1998new}. A comprehensive review of meshfree methods can be found in \cite{chen2017meshfree}.


It is noted that most of the above-mentioned FE and meshfree methods are formulated based on the Galerkin weak form. Therefore, for nonlinear mechanics analysis, the variational principle employed in these numerical methods has to be linearized through the Newton-Raphson method
to obtain incremental variational formulations that can be effectively solved by linear matrix solvers~\cite{holzapfel2002nonlinear,belytschko2014nonlinear}. 
In the incremental form, deriving consistent tangent stiffness is essential for numerical convergence but can be challenging when the complex constitutive equations have complicated forms.
In recent years, owing to the rapidly increasing computing power and the generalizable representation capabilities of deep neural networks (DNNs) \cite{hornik1991approximation}, significant advancements have been propelled in computational mechanics, ranging from neural-network based constitutive modeling \cite{vlassis2020geometric, linka2021constitutive, masi2021thermodynamics}, data-driven model-free computing \cite{kirchdoerfer2016data,he2020physics_lcdd,he2021manifold}, to data-driven multiscale modeling \cite{liu2019deep, masi2022multiscale}. In most of these studies, DNNs are employed as data-driven surrogates for representing material constitutive relations~\cite{fuhg2024review}.
Meanwhile, another emerging paradigm known as physics-informed machine learning (PIML)~\cite{karniadakis2021physics}, leveraging DNNs as function approximators for solving general partial differential equations (PDEs),
has also recently been proven effective in modeling computational mechanics problems~\cite{haghighat2021physics,samaniego2020energy,rao2021physics,zhang2022analyses,he2020physics,du2024neural}.

Among the spectrum of PIML approaches, Raissi et al. \cite{raissi2019physics} pioneered the development of physics-informed neural networks (PINNs), 
where the established physical laws are incorporated into the loss function through a residual-based minimization of the strong form of the governing equations, leading to enhanced training performance and reduced reliance on data measurements. 
Given the inherent adaptivity of the physics-informed framework, PINNs have been widely applied to various material modeling problems~\cite{haghighat2021physics,niu2023modeling,abueidda2021meshless}.


Nevertheless, a major challenge in using DNNs for nonlinear mechanics problems is that PINNs, solving PDEs by minimizing strong-form residuals, require a large number of sampling (collocation) points across the problem domain to achieve adequate solution resolution~\cite{abueidda2022deep,he2021physics,du2023modeling}.
Additionally, the presence of high-order derivatives in the residual-based loss function poses significant challenges to model training~\cite{krishnapriyan2021characterizing,samaniego2020energy,du2024neural}. 
For instance, in the elasticity problem, the requirement of second-order spatial gradients of the displacement field makes computations particularly expensive.
This difficulty has prompted researchers to explore mixed-variable output in PINNs, where both displacement and stress are independently approximated by DNN models~\cite{haghighat2021physics,rao2021physics,fuhg2022mixed,rezaei2022mixed}. 

Considering the same PDE can be expressed in different forms, alternative PIML approaches with reduced order of derivatives have been developed. These approaches include the energy form \cite{yu2018deep,samaniego2020energy,baek2022neural,khara2024neufenet}, 
Ritz form~\cite{yu2018deep}, 
weak form \cite{kharazmi2021hp,khodayi2020varnet,gao2022physics}, and meshfree local variational form~\cite{du2024neural}, which offer improved stability and convergence properties.
Drawing from the concept of energy-based PINNs, the deep energy method (DEM) has been applied to non-elastic materials such as hyperelasticity\cite{nguyen2020deep,abueidda2022deep} and elastoplasticity \cite{he2023deep}. 
However, it should be pointed out that since the loss function in DEM relies on the strain energy form, it may not be applicable to complex and non-conservative materials for which energy formulations are not well-defined.
On the other hand, the weak (variational) formulation, derived from Galerkin or Petrov-Galerkin methods, provides a flexible and consistent way to solve the governing equations weakly. In this way, the accuracy and requirement of continuity in solution depend on the selection of test functions~\cite{kharazmi2021hp,du2024neural}.

To establish a truly meshfree scheme that circumvents the need for conforming meshes in domain integration in energy and weak form-based PIML, 
Du and He \cite{du2024neural} recently proposed the neural-integrated meshfree (NIM) method based on the local Petrov-Galerkin formulation. Compared to the \textit{hq}-VPINN~\cite{kharazmi2021hp} and DEM, NIM allows the construction of loss functions over overlapping subdomains while maintaining variational consistency~\cite{du2024neural}.
Additionally, NIM features the neuro-partition of unity (NeuroPU) hybrid approximation scheme, which combines a set of PU shape functions, defined based on prior knowledge, with a DNN architecture.
This hybrid approximation, coupled with the local variational forms of governing equations, results in a highly accurate, efficient, and differentiable-programming solver for computational mechanics.

It is noteworthy that the NIM method, adopting a hybrid framework blending neural network models and numerical discretization, forms a novel differentiable-programming paradigm that seeks solutions directly through end-to-end training, similar to PINNs and its variants.
Differentiable programming, which allows an algorithm to be differentiated via automatic differentiation (AD)~\cite{baydin2018automatic},
plays a crucial role in machine learning for obtaining gradients required in optimization algorithms.
With the dramatic improvements in accuracy and efficiency afforded by GPU computing, the
differentiable programming-based methods for solving PDEs have recently attracted growing interest~\cite{innes2019differentiable}, 
especially those methods established on various discretized representations derived from traditional numerical schemes, such as FE, meshfree, and finite volume methods~\cite{bezgin2023jax, gasick2023isogeometric, xue2023jax, dong2023deepfem,du2024neural}.
They integrate numerical linear algebra and gradient operations within neural network architectures, enabling significant acceleration on specialized hardware and enhancing the efficiency of model training.
Among these emerging methods, NIM distinguishes itself by maintaining its meshfree properties, akin to PINNs. 
This characteristic not only preserves the flexibility inherent in meshfree methods but also leverages the robust differentiability of neural networks to advance the solution of complex PDEs in computational mechanics~\cite{du2024neural}.



This study aims to extend the NIM method to finite strain nonlinear material modeling by employing the consistent local variational formulation across different constitutive models. Unlike traditional numerical methods that require linearization to obtain consistent stiffness, we will demonstrate that the NIM solver can handle nonlinear constitutive models directly through differentiable programming. Additionally, by introducing a specialized PU shape function, we integrate the boundary singular kernel method \cite{chen2000new} to the NIM framework. Therefore, the essential boundary conditions (EBC) will be directly embedded into the NeuroPU approximation, eliminating the need for the penalty method used for EBC enforcement. The superior performance of the NIM method for forward modeling will first be demonstrated through detailed numerical tests on 1D and 2D homogeneous and isotropic hyperelasticity problems.

Furthermore, considering the critical importance of identifying heterogeneous micromechanical properties of hyperelastic materials,
such as biological tissues, for various medical and engineering applications, this study also explores the NIM method's potential for inverse modeling. 
To this end, in addition to the displacement approximation, an auxiliary NeuroPU model is adopted to approximate the unknown elastic modulus field. The NIM method then assimilates strain data measurements to estimate the hidden full-field heterogeneous elastic properties of hyperelastic materials. This application showcases the data-physics fusion capacity naturally inherent in NIM enabled by differential programming.

The main contributions of this study are summarized as follows:
\begin{itemize}
    \item  We introduce a novel meshfree PIML method constructed based on a consistent local variational framework for nonlinear material modeling without using material tangent stiffness.
    \item We incorporate the boundary singular kernel method into NeuroPU for effective EBC enforcement.
    \item We demonstrate NIM's effectiveness in the inverse identification of the heterogeneous parameter field in hyperelastic materials.
    \item We implement the NIM method using the JAX framework to enable GPU acceleration.
\end{itemize}

The remainder of the paper is organized as follows: Section \ref{sec:neuro} provides a review of the neuro-partition of unity (NeuroPU) approximation. In Section \ref{sec:nim}, we develop the NIM framework for hyperelasticity models.
Numerical tests for forward modeling and inverse modeling are presented in Section \ref{sec:result} and Section \ref{sec:result_inv}, respectively. 
Section \ref{sec:conclusion} concludes the paper by summarizing the main findings and contributions.

\section{Hybrid Approximation: NeuroPU}\label{sec:neuro}

Enabled by the ability to approximate arbitrary continuous functions according to the universal approximation theorem~\cite{hornik1991approximation, blum1991approximation}, DNNs have been increasingly used 
in physics-informed machine learning (PIML) methods for solving PDEs arising from engineering and science problems 
\cite{karniadakis2021physics, kashinath2021physics, wu2018physics, zobeiry2021physics}.
One notable PIML method is PINNs \cite{raissi2019physics}, which use DNNs with spatial and temporal coordinates as inputs to serve as an ansatz solution for the PDEs of interest. These DNNs are then optimized by minimizing the total residuals of the governing equations, initial/boundary conditions, and available measurements.

Nevertheless, there is growing interest in developing hybrid \textit{neuro-symbolic}  approximation approaches for solving PDEs, instead of solely using DNNs, to enhance the representational capacity and training efficiency of the network models \cite{zhang2021hierarchical,du2024neural,baek2022neural,fang2021high}. 
In this study, we introduce the recently proposed NeuroPU approximation \cite{du2024neural} for nonlinear material modeling, which involves interpolating the output values of a DNN model using a set of partition of unity (PU) basis functions. For completeness, the concept of NeuroPU will be summarized as follows. 

Without loss of generality, consider a generic PDE whose solution $u(\bm x)$ depends on a set of problem-specific parameters $\bm{\mu}$. Let a network $\mathcal{N}_{\theta}$ define a mapping from the system parameters $\bm{\mu}$ to a set of nodal solutions $\bm{\hat d} := \{ \hat d_I \}_{I=1}^{N_h}$ that characterize the discrete solution field of the PDE, where $N_h$ denotes the number of nodes associated with the underlying discretization.
Formally, the $\mathcal{N}_{\theta}$ maps an input space $\mathbb{R}^{c}$ to an output space $\mathbb{R}^{N_h}$, that is:
\begin{equation}\label{eq:nma_dnn}
\mathcal{N}_{\theta}(\bm{\mu}): \bm{\mu} \in \mathbb{R}^{c}  \mapsto \hat{\bm{d}} \in \mathbb{R}^{N_h}
\end{equation}
where the network $\mathcal{N}_{\theta}$ is parameterized by $\bm \theta$ (i.e., a collection of trainable weights and biases), and the output layer consists of $N_h$ components.


\begin{figure}[htb]
    \centering	\includegraphics[angle=0,width=0.75\textwidth]{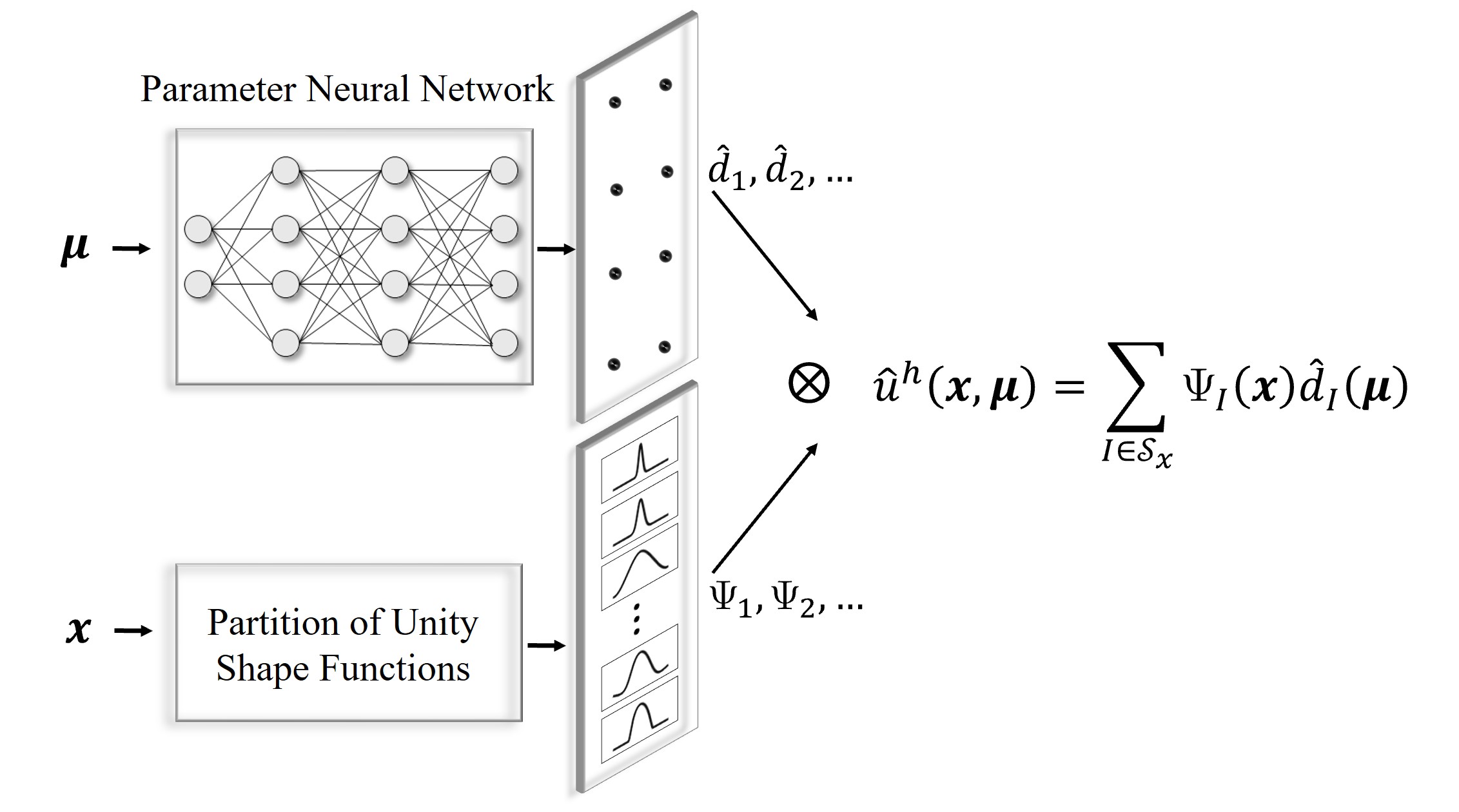}
    \caption{Schematic of neuro-partition of unity (NeuroPU) approximation for the solution $u(\bm x, \bm \mu)$, which is constructed based on the inner product of two function blocks: a neural network block and a block of symbolic PU functions.
    The neural network block takes the system parameters $\bm \mu$ as inputs, and outputs a set of nodal coefficient functions $\{\hat{d}_I\}_{I=1}^{N_h}$. The symbolic PU block produces a set of PU shape functions $\{\Psi_I\}_{I=1}^{N_h}$ that possess the PU condition and local compactness.
    } 
\label{fig:NeuroPU}
\end{figure}


As shown in Figure \ref{fig:NeuroPU}, the NeuroPU approximation, as a hybrid approach, combines 
the DNN represented nodal coefficient functions (Eq. \eqref{eq:nma_dnn})
and a set of PU shape functions $\{\Psi_I(\bm x)\}_{I=1}^{N_h}$ defined over the physical domain. 
Taking a parametric scalar field $u(\bm x, \bm \mu)$ as an example, its NeuroPU approximation is  expressed as
\begin{equation}
\hat{u}^h(\boldsymbol{x},\boldsymbol{\mu};\bm \theta)=\sum_{I \in \mathcal{S}_x} \Psi_I(\boldsymbol{x}) \hat{d}_I (\bm{\mu}; \bm \theta),
\label{eq:nmm}
\end{equation} 
where $\hat{u}^h$ represents the approximated solution,
$\mathcal{S}_x$ represents the set of nodes contributing to the interpolation at $\bm{x}$,
$\{\Psi_I\}_{I=1}^{N_h}$ denote the predefined PU shape functions, 
and $\hat{{d}}_I$ stands for the corresponding $I$th nodal coefficient output of 
the neural network $\mathcal{N}_{\theta}$.
The hat symbol $\hat{.}$ denotes a variable approximated by a DNN model.
For the sake of brevity, we omit the symbols $\bm{\mu}$ and $\bm{\theta}$ in the following expressions.

Similar to the study \cite{du2024neural},
we choose to use the reproducing kernel (RK) shape function \cite{liu1995reproducing} for the NeuroPU approximation due to its well-controlled smoothness and locality. 
The construction of RK shape functions is provided in \ref{sec:RKPM}.
However, it is worthwhile to point out that any type of PU shape function \cite{lu2023convolution}, such as finite element (FE) shape functions, radial basis interpolation, 
and non-uniform rational B-spline (NURBS) basis functions
, is also applicable to the NeuroPU scheme.



By using this hybrid NeuroPU approximation, the spatial coordinates are only involved in the shape functions. Thus, the spatial derivatives of Eq.  \eqref{eq:nmm} can be expressed as 
\begin{equation}\label{eq:grad_nma}
\nabla \hat{u}^h(\boldsymbol{x})=\sum_{I \in \mathcal{S}_x} \nabla \Psi_{I}(\boldsymbol{x}) \hat{d}_I,
\end{equation} 
It is noted that the spatial gradients of shape functions $\nabla \Psi_{I}$ can be precomputed  \textit{off-line}, rather than using automatic differentiation (AD) \cite{baydin2018automatic}. This approach significantly reduces computational overhead during the training process due to the decreased number of AD operations. 
Furthermore, NeuroPU can mitigate the training difficulties of DNNs
because it avoids the high-order derivatives that commonly require AD in constructing loss functions for strong-form, weak-form, and energy-form PINN methods \cite{krishnapriyan2021characterizing,kharazmi2021hp,nguyen2020deep,du2024neural}.

Thanks to the reduction in the order of derivatives, our numerical experiments also indicate that the proposed NIM can achieve satisfactory displacement solution without using the mixed variables scheme, which is commonly employed in PINNs and DEM to guarantee simulation accuracy when dealing with solid material modeling~\cite{rao2021physics,fuhg2022mixed,rezaei2022mixed}.  


Like other spatial approximation methods, such as Lagrange polynomials and spectral method, NeuroPU can serve as a general approximation tool for arbitrary functions, not just the solution field.
For example, in Section \ref{sec:result_inv}, the NeuroPU approach is also applied to approximating the heterogeneous material property field. The solution procedure with multiple NeuroPU models will be detailed in Section \ref{sec:result_inv}.



\emph{Remark 2.1.}
The concept of partition of unity (PU) functions has been demonstrated to be useful in deep learning-based methods \cite{zhang2021hierarchical, park2023convolution, baek2022neural, baek2024neural,lee2021partition}. In hierarchical deep-learning neural networks (HiDeNN)~\cite{zhang2021hierarchical, park2023convolution}, the structured DNN blocks are used to construct PU shape functions and solve  PDEs by training the DNNs.
Similar block-level NN approximation is  extended in \cite{baek2022neural, baek2024neural} for modeling localization and fractures.
In this approach, PU shape functions are adopted to handle the smooth part of the solution space, while the NN blocks, pre-trained offline, capture the non-smooth solution. 
Although the proposed NeuroPU resembles other PU approximation approaches in ensuring the PU property, it is established in
a general setting related to DeepONet \cite{lu2021learning}, as shown in Figure \ref{fig:NeuroPU}, where the parameter representation space ("branch net") and the basis function space ("trunk net") are combined via inner product. 
Unlike the original DeepONet model, pre-defined PU shape functions are constructed to replace the 
the learnable trunk net in NeuroPU, enhancing training efficiency and providing controllable accuracy.
\section{Neural-Integrated Meshfree (NIM) Nonlinear Modeling Framework}\label{sec:nim}

In this section, the NIM framework will be extended for modeling nonlinear materials, where the NeuroPU approach is introduced to approximate the displacement solution, and a local variational formulation based on the Petrov–Galerkin method \cite{du2024neural} is developed for simulating inelastic solids at finite-strain. Several homogeneous and isotropic hyperelastic material models, such as the Neo-Hookean model, are considered for demonstration.




\subsection{Problem formulation for hyperelasticity}\label{sec:pro_hyper}

Let the $n$-dimensional problem domain under the initial/undeformed configuration be denoted by $\Omega^X  \in \mathbb{R}^{n}$.
The displacement field is defined as $\boldsymbol{u} = \boldsymbol{u} (\bm{X})$, leading a material particle at position $\boldsymbol{X} \in \Omega^X$ to $\boldsymbol{x} \in \Omega^x$, where $\bm x = \bm X + \bm u (\bm{X})$, and $\Omega^x$ represents the current configuration.

The \textit{total Lagrangian formulation} is commonly employed in simulations of deformation involving nonlinear elastic materials. 
The equations governing the deformation of a solid in the undeformed configuration are given as:
\begin{equation} \label{eq:hyperelastic}
\begin{cases}
\nabla_{X} \cdot \boldsymbol{P}+\boldsymbol{f}= \bm 0, \quad & \text{in} \quad \Omega^X \\ 
\boldsymbol{n} \cdot \boldsymbol{P}=\overline{\boldsymbol{t}}, \quad & \text{on} \quad \Gamma_t^X \\
\bm u ={\overline{\bm u}}, \quad & \text{on} \quad \Gamma_g^X
\end{cases}
\end{equation}
where $\Gamma_g^X$ and $\Gamma_t^X$ denote the boundaries where essential (EBCs) and natural boundary conditions (NBCs) are respectively prescribed, with satisfying $\Gamma^X = \Gamma_g^X \cup \Gamma_t^X$ and $\Gamma_g^X \cap \Gamma_t^X = \emptyset $,
$\nabla_{X}$ is the gradient operator applied on the initial configuration, $\bm f$ is the body force, $\overline{\bm u}$ and $ \overline{\bm t}$ are the displacement and traction values prescribed on $\Gamma_g^X $ and $\Gamma_t^X$, respectively, and $\bm n$ is the surface normal on $\Gamma_t^X$. $\bm P$ is the $1^{st}$ Piola-Kirchhoff stress. 
For simplicity in the exposition of methodology derivation, the superscript $^X$ associated with domain and boundary symbols will be omitted in the subsequent discussion.

For hyperelastic materials, the stress-strain constitutive relation is derived from the strain-energy density potential that characterizes the material model:
\begin{equation} \label{eq:PK}
\boldsymbol{P}=\frac{\partial W}{\partial \boldsymbol{F}}
\end{equation}
where $\bm F$ is the deformation gradient defined as $\boldsymbol{F}=\nabla_X \bm u(\boldsymbol{X}) + \bm I$ with $\bm I$ being the identity matrix, and $W$ is the strain energy density function that is usually defined as a function of the deformation invariant.



 
For completeness, the hyperelastic material models considered in this study (see Section \ref{sec:result}) are described as follows:
\begin{equation} \label{eq:psi_st}
\text{St. Venant-Kirchhoff:   } W=\frac{1}{2}(\lambda+2 \mu) I_1^2-2 \mu I_2
\end{equation}
and
\begin{equation} \label{eq:psi_nh}
\text{Neo-Hookean:   } W=\frac{1}{2} \lambda[\log (J)]^2-\mu \log (J)+\frac{1}{2} \mu\left(I_1-3\right)
\end{equation}
where $I_1=\operatorname{trace}(\boldsymbol{C})$, $I_2=\frac{1}{2} [\operatorname{trace}(\boldsymbol{C})^2 - \operatorname{trace}(\boldsymbol{C}^2)]$, and $J$ is the elastic volume ratio defined as $J=\operatorname{det}(\boldsymbol{F}) = \sqrt{I_3}$.
Here, $I_k$'s ($k=1,2,3$) are the first, second, and third invariants of the right Cauchy-Green deformation tensor $\boldsymbol{C}=\boldsymbol{F}^T \cdot \boldsymbol{F}$.
$\lambda$ and $\mu$ are Lame parameters, which are related to Young's modulus $E$ and Poisson's ratio $\nu$ as follows
\begin{equation}\label{eq:lame}
\lambda=\frac{E \nu}{(1+\nu)(1-2 \nu)}, \quad \mu=\frac{E}{2(1+\nu)}
\end{equation}

\subsection{NIM nonlinear modeling based on local variational formulation}\label{sec:formulation}
To solve the hyperelasticity problem described above using the NIM method, we first need to approximate the displacement field. This is achieved using the NeuroPU approach introduced in Section \ref{sec:neuro}, as detailed below. The approximation of displacement vector, $\hat{\bm u}^h \in \mathbb{R}^{n}$ ($n=1,2$ or $3$), in the initial configuration is given as
\begin{equation}
\hat{\bm u}^h(\boldsymbol{X})=\sum_{I \in \mathcal{S}_X} \Psi_I(\boldsymbol{X}) \hat{\bm d}_I, \quad \bm X \in \Omega
\label{eq:u_hyper}
\end{equation} 
where $\hat{\bm d}_I = [ \hat{ d}_I^1, \hat{ d}_I^2, \ldots, \hat{ d}_I^n ]$ is the nodal displacement vector associated with the $I$th node, and each component $\hat{d}_I^j$, for $j = 1,\ldots,n$, represents the nodal displacement value along the $j$th direction.
In NeuroPU construction, $ \hat{ d}_I^j$ is obtained from the $I$th nodal output of the $j$th DNN model $\mathcal{N}_{\theta}^j$ (or the $j$th channel of $\mathcal{N}_{\theta}$ if an $n$-channel convolutional neural network structure is used).
Subsequently, the NeuroPU approximated deformation gradient tensor, $\hat{\bm F}^h(\bm X)$, is given as
\begin{equation} \label{eq:F_hyper}
\begin{aligned}
\hat{\bm F}^h(\bm X)&=\nabla_X \bm \hat{\bm u}^h(\bm{X}) + \bm I
            =\sum_{I \in \mathcal{S}_X} \nabla_X \Psi_I(\boldsymbol{X}) \hat{\bm d}_I + \bm I
\end{aligned}
\end{equation}
 With $\hat{\bm F}^h$ computed by the NeuroPU model, the approximated strain energy density $\hat W^h$ can be easily obtained following its mathematical formulation, such as Eqs. \eqref{eq:psi_st} and \eqref{eq:psi_nh},
 which also leads to the approximated $1^{st}$ Piola-Kirchhoff stress $\hat {\bm P}^h$ in \eqref{eq:PK} as follows
\begin{equation}
\hat {\bm P}^h = \frac{\partial \hat W^h}{\partial \hat {\bm F}^h}
\label{eq:P_approximate}
\end{equation}
Since the NeuroPU model is implemented in a differentiable programming manner (see Section \ref{sec:jax}), the involved derivatives, e.g., Eq. \eqref{eq:P_approximate}, can be efficiently handled via automatic differentiation and backpropagation during training, circumventing the need of symbolic or numerical differentiation.

With the displacement approximation defined above, the NIM modeling framework for generic finite-strain inelastic materials can thus be developed.
It should be noted that in NIM, the \textit{local variational formulation} will be incorporated as part of the solution procedure.

Let $\mathcal{T}$ be defined as a set of local subdomains $\{ \Omega_s \}_{s=1}^{N_{\mathcal{T}}}$, where  $N_{\mathcal{T}} = |\mathcal{T}|$ represents the number of subdomains distributed over the reference domain $\Omega$. Ensure that their union covers the whole domain, i.e., $\bar{\Omega} \subseteq \bigcup_{s \in \mathcal{T}} \Omega_s$.
To derive the local variational formulation, we introduce $N_v$ compactly supported test functions $\{ \bm{v}^{(k)} \}_{k=1}^{N_v}$ for each subdomain $\Omega_s$. Generally, $N_v$ is chosen to be at least $n$ (i.e., $N_v\geqslant n$) to prevent the rank deficiency \cite{atluri1998new,atluri2000new}.

By substituting the approximated stress from \eqref{eq:P_approximate} into the equilibrium equation in \eqref{eq:hyperelastic} and multiplying by the $k$th test function over $\Omega_s$, we obtain the local weighted residual $\bar{\mathcal{R}}_{s}^{(k)}$ as follows

\begin{equation}
\bar{\mathcal{R}}_{s}^{(k)} = \int_{\Omega_s} {\bm v^{(k)}} \cdot\left(\nabla_{\bm X} \cdot \hat {\bm P}^h+\boldsymbol{f}\right) d \Omega
\label{eq:weighted}
\end{equation}
As discussed in \cite{atluri2000new,du2024neural}, $N_v = n$ is adopted in our study, and $\bm{v}^{(k)}(\bm X)$ can be simply chosen as ${v}^{(k)}_i=v(\bm X)\delta_{ik}$ with $v(\bm X)$ being scalar function defined over subdomains.


Applying integration by parts and divergence theorem to the weighted residual forms  \eqref{eq:weighted} yields the following local variational residuals
\begin{equation}
\begin{aligned}
\mathcal{R}_{s}^{(k)} &= \int_{\Omega_s} \nabla_{X} \boldsymbol{v}^{(k)}: \hat {\bm P}^h d \Omega 
- \int_{\Omega_s} \boldsymbol{v}^{(k)} \cdot \boldsymbol{f} d \Omega \\
&- \int_{L_s} \boldsymbol{v}^{(k)} \cdot \hat {\bm t}^h d \Gamma
- \int_{\Gamma_{s g}} \boldsymbol{v}^{(k)} \cdot \hat {\bm t}^h d \Gamma 
 - \int_{\Gamma_{s t}} \boldsymbol{v}^{(k)} \cdot \overline{\boldsymbol{t}} d \Gamma 
\end{aligned}
\label{eq:local_inte}
\end{equation}
where the traction is given by $\hat {\bm t}^h= \hat {\bm P}^h \cdot \bm{n}$. The local boundary $\partial \Omega_s$ is divided into  $\Gamma_s$ and $L_s$, which are located on the global boundary $\Gamma = \partial \Omega$ and within the domain $\Omega$, respectively, i.e., $\Gamma_s = \partial \Omega_s \cap \Gamma$, and $L_s = \partial \Omega_s \setminus \Gamma$. Specifically, $\Gamma_s$ is further divided into $\Gamma_{s g}$ and $\Gamma_{s t}$, representing the local boundary segments on which the EBCs and NBCs are specified, respectively.
We can observe from Eq. \eqref{eq:local_inte} that, this developed local variational setting allows the use of overlapping subdomains for NIM modeling, ensuring its meshfree property.

Following \cite{du2024neural}, the residual form Eq. \eqref{eq:local_inte} can be further simplified by selecting test functions $v(\bm X)$ that possesses special
properties. For example, if using the Heaviside step function as test function
    \begin{equation}
    v(\boldsymbol{X})= \begin{cases}0 & \boldsymbol{X} \notin \left(\Omega_s \cup L_s\right) \\ 1 & \boldsymbol{X} \in \left(\Omega_s \cup L_s\right)\end{cases}
    \label{eq:heavi}
    \end{equation}
the local variational residual is rewritten as    
\begin{equation}
\begin{aligned}
\text{NIM/h}: \quad 
\mathcal{R}_{s}^{(k)} =  
\int_{\Omega_s} \boldsymbol{v}^{(k)} \cdot \boldsymbol{f} d \Omega
+
\int_{L_s} \boldsymbol{v}^{(k)} \cdot (\hat {\bm P}^h \cdot \bm{n}) d \Gamma 
\\
+ \int_{\Gamma_{s g}} \boldsymbol{v}^{(k)} \cdot (\hat {\bm P}^h \cdot \bm{n}) d \Gamma   + \int_{\Gamma_{s t}} \boldsymbol{v}^{(k)} \cdot \overline{\boldsymbol{t}} d \Gamma 
\end{aligned}
\label{eq:local_weak_h}
\end{equation}
where $\boldsymbol{v} = [\boldsymbol{v}^{(1)}, \cdots, \boldsymbol{v}^{(n)}] = v(\bm X) \bm I$, with $\bm I$ being an identity tensor.
Accordingly, the NIM solver that employs the Heaviside step function is referred to as NIM/h.

Alternatively, to enhance approximation accuracy, we can employ smooth high-order test functions that vanish on local boundaries $\partial \Omega_s$.
For example, consider the cubic B-spline function (see Eq. \eqref{eq:kernel_B}) defined on a 2D rectangle or circle subdomain, leading to
the following residual formulation, termed NIM/c:
\begin{equation}
\begin{aligned}
\text{NIM/c}: \quad 
\mathcal{R}_{s}^{(k)} = 
\int_{\Omega_s} \nabla_{\bm X} \boldsymbol{v}^{(k)}: \hat {\bm P}^h d \Omega 
- \int_{\Omega_s} \boldsymbol{v}^{(k)} \cdot \boldsymbol{f} d \Omega
\\
- \int_{\Gamma_{s g}} \boldsymbol{v}^{(k)} \cdot (\hat {\bm P}^h \cdot \bm{n}) d \Gamma 
- \int_{\Gamma_{s t}} \boldsymbol{v}^{(k)} \cdot \overline{\boldsymbol{t}} d \Gamma 
\end{aligned}
\label{eq:local_weak_c}
\end{equation}

Compared to the general NIM residual formulation in Eq. \eqref{eq:local_inte}, we note that NIM/h \eqref{eq:local_weak_h} eliminates the domain integral of strain energy over $\Omega_s$,
whereas the the boundary integral on $L_s$ disappears in NIM/c \eqref{eq:local_weak_c}.

Integrating the local variational residuals over the whole set of subdomains $ \{\Omega_s\}_{s=1}^{N_\mathcal{T}}$ and considering $N_v$ arbitrary local test functions, the total loss function of the NIM method is written as:

\begin{equation}
\begin{aligned}
\mathcal{L}(\bm \theta)&=\frac{1}{N_\mathcal{T}}\sum_{k=1}^{N_v}\sum_{s=1}^{N_\mathcal{T}}\left\|{\mathcal{R}}_s^{(k)} \right\|^2 
\label{eq:loss_v}
\end{aligned}
\end{equation}
Here, it is noted that, different from the strong-form PINNs, the traction boundary condition is naturally incorporated into the local variational residual formulation, as shown in Eqs. \eqref{eq:local_weak_h} and \eqref{eq:local_weak_c}, so that NIM circumvents the necessity of adding an penalty term corresponding to natural boundary conditions in the loss function.
On the other hand, we will specifically discuss the imposition of EBCs in Section \ref{sec:singular}.

\begin{figure}[htb]
	\centering
	\includegraphics[angle=0,width=0.8\textwidth]{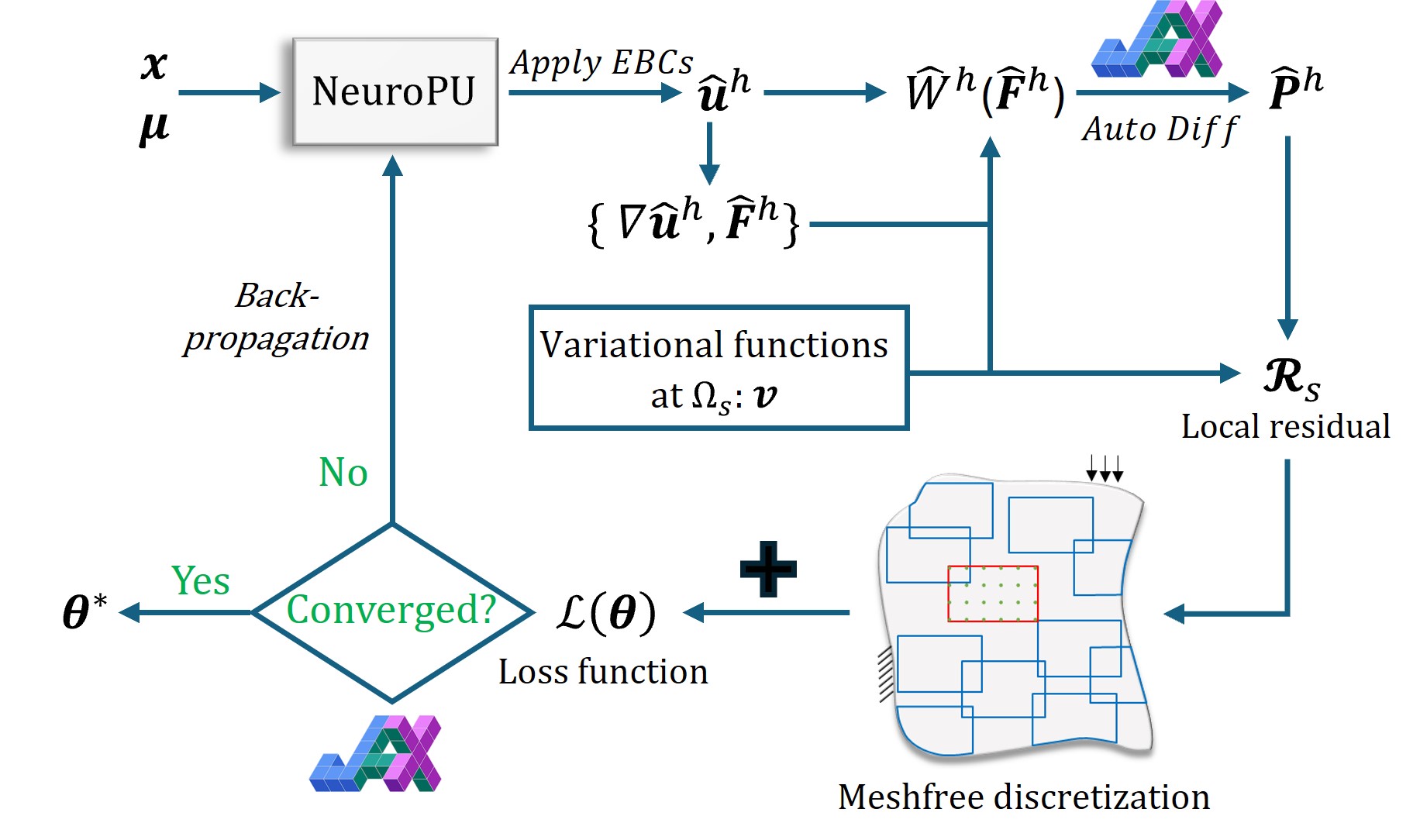}
 \caption{The workflow of the neural-integrated meshfree (NIM) nonlinear modeling framework for hyperelastic materials, where the strain energy density $\hat{W}^h$ is modeled as a function of the displacement $\hat{\bm u}^h$ and its derivatives, approximated using the NeuroPU approach. The PK stress $\hat {\bm P}^h = \frac{\partial \hat W^h}{\partial \hat {\bm F}^h}$ is computed through automatic differentiation provided by JAX. 
Within the meshfree discretization, the red box indicates one of the subdomains $\Omega_s$, which is associated with a local variational residual term $ \bm {\mathcal R}_s$ (Eq. \eqref{eq:local_inte}). 
 The summation of $\bm{\mathcal R}_s$ over the subdomains $\{\Omega_s\}_{s=1}^{N_{\mathcal{T}}}$ constitutes the total loss function $\mathcal{L}(\boldsymbol{\theta})$.}
\label{fig:nim}
\end{figure}

Figure \ref{fig:nim} illustrates the workflow of the NIM method for hyperelastic material modeling, where the collection of trainable parameters $\bm \theta$ of the NeuroPU approximation for the displacement field $\bm {\hat u}^h$ is updated through minimizing the loss function $\mathcal{L}(\bm \theta)$ in \eqref{eq:loss_v}. 
Upon convergence, the displacement solution can be obtained by using \eqref{eq:u_hyper} with the optimized parameters $\bm \theta^*$.


\emph{Remark 3.1.}
It should be emphasized that when using conventional numerical methods, such as FEM, for nonlinear material modeling \cite{wriggers2008nonlinear,belytschko2014nonlinear}, the Newton-Raphson method is employed to solve the nonlinear equation iteratively. Thus, a tangent stiffness matrix, derived from the Fréchet derivative of the weak-form residual, is required for iteration updates.
However, the proposed differentiable NIM nonlinear modeling framework eliminates the need to use the tangent stiffness matrix, allowing the displacement solution to be obtained by optimizing the loss function \eqref{eq:loss_v}, which is constructed simply from the local variational residual without its derivatives.


\subsection{EBC enforcement: Boundary singular kernel method}\label{sec:singular}
While the local variational formulation naturally incorporates NBCs, the imposition of EBCs requires special treatment. In this section, we introduce the boundary singular kernel method into the NIM framework for enforcing EBCs. 

Various different approaches have been proposed to impose EBCs in PIML, generally categorized into the "soft" (or weak) \cite{raissi2019physics,he2021physics} and "hard" approaches \cite{lagaris1998artificial,berg2018unified,SUKUMAR2022114333}. 
The soft approach involves penalizing the deviations associated with initial and boundary conditions and adding these penalization terms, scaled with proper weight coefficients, into the loss function.
On the other hand, the hard approach usually relies on
modifying the network architecture or introducing custom distance functions \cite{berg2018unified} within the DNN ansatz such that the boundary conditions are satisfied automatically, resulting in hard-constraint methods.

In the linear version of NIM \cite{du2024neural}, the EBCs are enforced via the penalty method. 
Nevertheless,
owing to the flexibility of customizing basis functions in the hybrid NeuroPU approximation,
we propose a new method based on the boundary singular kernel method \cite{chen2000new} for NIM modeling, 
which offers a straightforward way to strongly impose EBCs
without the use of the penalty terms. 



As presented in \ref{sec:RKPM}, the standard RK shape functions $\{\Psi_I\}_{I=1}^{N_h}$ are constructed by imposing the reproducing conditions on a set of kernel functions $\phi_a(\bm{X}_I - \bm {X})$. 
Invoking the boundary singular kernel method \cite{chen2000new},
a singularity will be introduced to the kernel functions associated with the nodes located on the essential boundary, such that for $\mathcal{S}_{sk}= \{I | \bm{X}_I \in \Gamma_g, I = 1, \cdots, N_h\}$, the resulting singular kernel shape functions 
$ \{ \tilde \Psi_I \}_{I \in \mathcal{S}_{sk}} $ 
have the following property:
\begin{equation}
\tilde{\Psi}_I(\bm X \rightarrow \bm{X}_I) = 1
\label{eq:delta_1}
\end{equation}
Due to the PU condition as well as the reproducing properties, other unmodified RK shape functions $ \{\Psi_J \}$, for $ J \notin \mathcal{S}_{sk}$, ensure
\begin{equation}
\Psi_J(\bm X \rightarrow \bm{X}_I) = 0
\label{eq:delta_2}
\end{equation}
However, it is important to note that since $\tilde{\Psi}_I(\bm X \rightarrow \bm{X}_J) \neq 0$, $\tilde{\Psi}_I$ is not an interpolation function that possesses Kronecker delta property. The details on constructing the singular kernel shape functions can be found in \cite{chen2000new}.

By integrating the singular kernel shape functions into the NeuroPU approximation, Eq. \eqref{eq:u_hyper} can be rewritten as
\begin{equation}
\hat {\boldsymbol u}^h(\boldsymbol{X})= \sum^{N_h}_{J =1 , J \notin \mathcal{S}_{sk}} \Psi_J(\boldsymbol{X}) \hat{\bm d}_J 
+ \sum_{I \in \mathcal{S}_{sk}} \tilde \Psi_I(\boldsymbol{X}) \hat{\bm d}_I 
\label{eq:u_hyper_singular}
\end{equation} 
Given the properties in Eqs. \eqref{eq:delta_1} and \eqref{eq:delta_2}, it shows
\begin{equation}
\bm{\hat u}^h(\bm{X}_I)=\hat{\bm d}_I \quad \text{for } I \in \mathcal{S}_{sk}
\end{equation} 
which means that the nodal coefficients $\hat{\bm d}_I$ for $I \in \mathcal{S}_{sk}$, represent the values of the approximated function $\bm{\hat u}^h$ at these nodes, 
despite the fact that $\tilde{\Psi}_I$ are not exactly interpolatory. 
This enables the direct imposition of boundary constraints by specifying the given boundary displacements to the DNN output components $\hat{\bm d}_I$ associated with the nodes on the essential boundary $\bm X_I \in \Gamma_g$, i.e., $\hat{d}_I^j := u_j(\bm X_I)$. 



\subsection{JAX implementation for NIM method}\label{sec:jax}
The commonly used Python-based platforms for implementing the differentiable programming framework include TensorFlow \cite{abadi2016tensorflow}, JAX \cite{jax2018github}, and PyTorch \cite{paszke2019pytorch}. 
As JAX offers explicit support for GPU-accelerated NumPy operations, just-in-time (JIT) compilation, and control over vectorization and parallel computing, we have chosen
JAX as the foundational framework for the developed NIM method to fully leverage GPU acceleration.

 Our methodology leverages \lstinline|jax.numpy| module to handle data exclusively in array format. This aligns with the array programming paradigm enhanced by JAX, supporting efficient data manipulation and computation. As illustrated in Table \ref{fig:app_gradient}, we consolidate shape function data from all quadrature points into single arrays \lstinline|DPHIX_all| and \lstinline|DPHIY_all|, and then multiply these arrays with the nodal coefficients \lstinline|d_x| and \lstinline|d_y|. Furthermore, \lstinline|jax.vmap| facilitates the vectorization of differentiation and array operations, as demonstrated in Table \ref{fig:pk_stress}, where \lstinline|jax.grad| is utilized for automatic differentiation. This not only reduces computational overhead but also eliminates the need for explicit loops over subdomains, which are conventionally a significant source of inefficiency.

\begin{table}[htb]
\centering
\SaveVerb{JAX}|JAX|

\centering
\begin{minted}[
frame=lines,
framesep=2mm,
fontsize=\footnotesize,
]{python}
Fxx = (DPHIX_all * dx[Index_all]).sum(axis = 1) + 1
Fxy = (DPHIY_all * dx[Index_all]).sum(axis = 1) + 0
Fyx = (DPHIX_all * dy[Index_all]).sum(axis = 1) + 0
Fyy = (DPHIY_all * dy[Index_all]).sum(axis = 1) + 1
\end{minted}
\vspace{-12pt}
\caption{JAX code snippet for approximation of deformation gradient.}
\label{fig:app_gradient}
\end{table}

\begin{table}[htb]
\centering
\SaveVerb{JAX}|JAX|

\centering
\begin{minted}[
frame=lines,
framesep=2mm,
fontsize=\footnotesize,
]{python}
Pxx = vmap(grad(self.SED, argnums=0), in_axes=(0, 0, 0, 0))(Fxx, Fxy, Fyx, Fyy)
Pxy = vmap(grad(self.SED, argnums=1), in_axes=(0, 0, 0, 0))(Fxx, Fxy, Fyx, Fyy)
Pyx = vmap(grad(self.SED, argnums=2), in_axes=(0, 0, 0, 0))(Fxx, Fxy, Fyx, Fyy)
Pyy = vmap(grad(self.SED, argnums=3), in_axes=(0, 0, 0, 0))(Fxx, Fxy, Fyx, Fyy)
\end{minted}
\vspace{-12pt}
\caption{JAX code snippet for approximation of $1^{st}$ Piola-Kirchhoff stress.}
\label{fig:pk_stress}
\end{table}

Additionally, \lstinline|jax.jit| compilation can be utilized to significantly accelerate both the training and evaluation phases of the NIM model.  By compiling critical sections of our code, such as the L-BFGS-B training optimizer (detailed in Table \ref{fig:BFGS}), we ensure that repetitive computations are executed more rapidly, thereby enhancing overall computational efficiency and scalability.

\begin{table}[htb]
\centering
\SaveVerb{JAX}|JAX|

\centering
\begin{minted}[
frame=lines,
framesep=2mm,
fontsize=\footnotesize,
]{python}
self.optimizer = jaxopt.ScipyMinimize(fun = self.loss,
    method = 'L-BFGS-B', maxiter = 50000, callback = self.callback, jit = True,
    options = {'maxfun': 50000, 'maxcor': 100,'maxls': 100, 
               'ftol': 1.0e-14, 'gtol': 1.0e-14})
\end{minted}
\vspace{-12pt}
\caption{JAX code snippet for the L-BFGS-B optimizer definition.}
\label{fig:BFGS}
\end{table}

The details of JAX-based implementation of the NIM method are available in the tutorial sample code: \url{https://github.com/IntelligentMechanicsLab/NIM-Tutorial}.

\section{Numerical Results: Forward Modeling}\label{sec:result}
In this section, we investigate the performance of the NIM method on various hyperelastic material modeling problems.
In addition, to demonstrate the unified capacities of predictive modeling and data assimilation provided by the proposed differentiable scheme, we will extend the current framework for inverse modeling of heterogeneous material properties, which will be presented in Section \ref{sec:result_inv}.

As discussed in Section \ref{sec:formulation}, the presented NIM framework offers the flexibility to select different \textit{test functions} $\bm{v}(\bm x)$ for constructing the local variational residuals. Specifically, we examine two typical test functions: the Heaviside step function and cubic B-spline function. Following the notation used in \cite{du2024neural}, the resultant NIM formulations are denoted as $\text{NIM/h}$ and $\text{NIM/c}$, respectively, as shown in Eqs. \eqref{eq:local_weak_h} and \eqref{eq:local_weak_c}.
Moreover, each subdomain is defined as a rectangle for 2D problems (or a line for 1D problems) with a side length of $2r=2\bar r h$, where $h$ is the characteristic nodal distance, and $\bar r$ is the normalized size of subdomains. To ensure sufficient integration accuracy, each subdomain is uniformly divided into $4 \times4$ segments for 2D (or 4 segments for 1D), where $5$ Gauss quadrature points per direction are allocated.

In terms of the \textit{trial functions} for the variational formulation, the quadratic meshfree shape functions (see \ref{sec:RKPM}) with normalized support size $\bar a = a/h=2.5$ are adopted for the NeuroPU approximation of displacement \eqref{eq:u_hyper}.
On the other hand, the neural network block in the NeuroPU approximation (Figure \ref{fig:nim}) adopts a network configuration with one hidden layer containing 10 neurons, denoted as $1 \times [10]$. 
The input to the neural network is set to an arbitrary constant, while the output size corresponds to the underlying discretization $N_h$. 
In all examples, the boundary singular kernel method outlined in Section \ref{sec:singular}
is employed  
for the imposition of EBCs, so that the corresponding RK shape functions are replaced by the singular kernel shape functions.

The L-BFGS-B optimizer from Scipy's built-in JAX library is utilized for training, with the settings configured as follows: $ftol = 1 \times 10^{-14}$, $gtol = 1 \times 10^{-14}$, $Iter \leq 5 \times 10^4$, $maxfun = 5 \times 10^4$, $maxcor=100$, and $maxls = 100$. The training hyperparameters are also shown in  Table \ref{fig:BFGS}.

For the evaluation of the performance, the relative $L_2$ error norm for the displacement vector is defined as follows
\begin{equation}
e_{L_2}=\frac{\| \hat{\bm u}^h-\bm u_{\text {ref}}\|}{\|\bm u_{\text {ref}}\|}
\label{eq:l2err}
\end{equation}
where $\bm u_{\text {ref}}$ denotes the reference solution. Unless otherwise stated, the FEM reference solutions are obtained using FEniCS \cite{alnaes2015fenics}.

\subsection{Uniaxial tension of hyperelastic material}\label{sec:hyper}

\begin{figure}[htb]
	\centering
	\includegraphics[angle=0,width=0.65\textwidth]{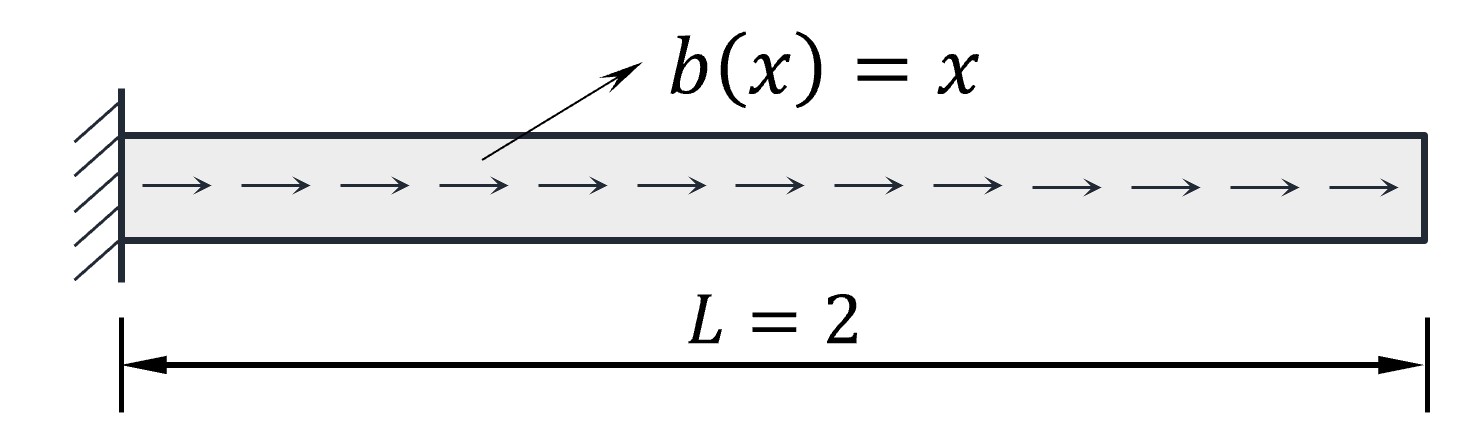}
	\caption{Illustration of the 1D bar problem subjected to an uniaxial tension.}
\label{fig:1dbar}
\end{figure}

We first consider a 1D problem \cite{nguyen2020deep} where a hyperelastic bar is under uniaxial tension with a body force $b(x)=x$, as shown in Figure \ref{fig:1dbar}. The strain energy function is given as
\begin{equation}
\Psi(\epsilon)=(1+\epsilon)^{\frac{3}{2}}-\frac{3}{2} \epsilon-1
\label{eq:1d_potential}
\end{equation}
where $\epsilon = du/dx $ denotes the gradient of displacement. The associated analytical solutions \cite{nguyen2020deep} of displacement and its gradient in $\Omega =[-1,1]$ are given as
\begin{equation}
\begin{cases}
u_{\text {ext }}(x)=\frac{1}{135} \left(68+105 x-40 x^3+3 x^5\right) \\ \epsilon_{\text {ext }}(x)=\frac{1}{9} \left(x^4-8 x^2+7\right)
\end{cases}
\end{equation}

Both NIM/h and NIM/c are applied to modeling this problem, where a uniform discretization of $N_h=41$ nodes is used for the NeuroPU approximation of displacement.
$N_{\mathcal{T}} = 101$ uniformly distributed subdomains with a normalized size $\bar r=2.5$ are utilized to construct the local variational formulation. 
By default, the network architecture of hidden layers for the NIM solvers is $1 \times [10]$ with the output layer of $N_h = 41$ neurons. 

\begin{figure}[htb]
	\centering
	\includegraphics[angle=0,width=0.8\textwidth]{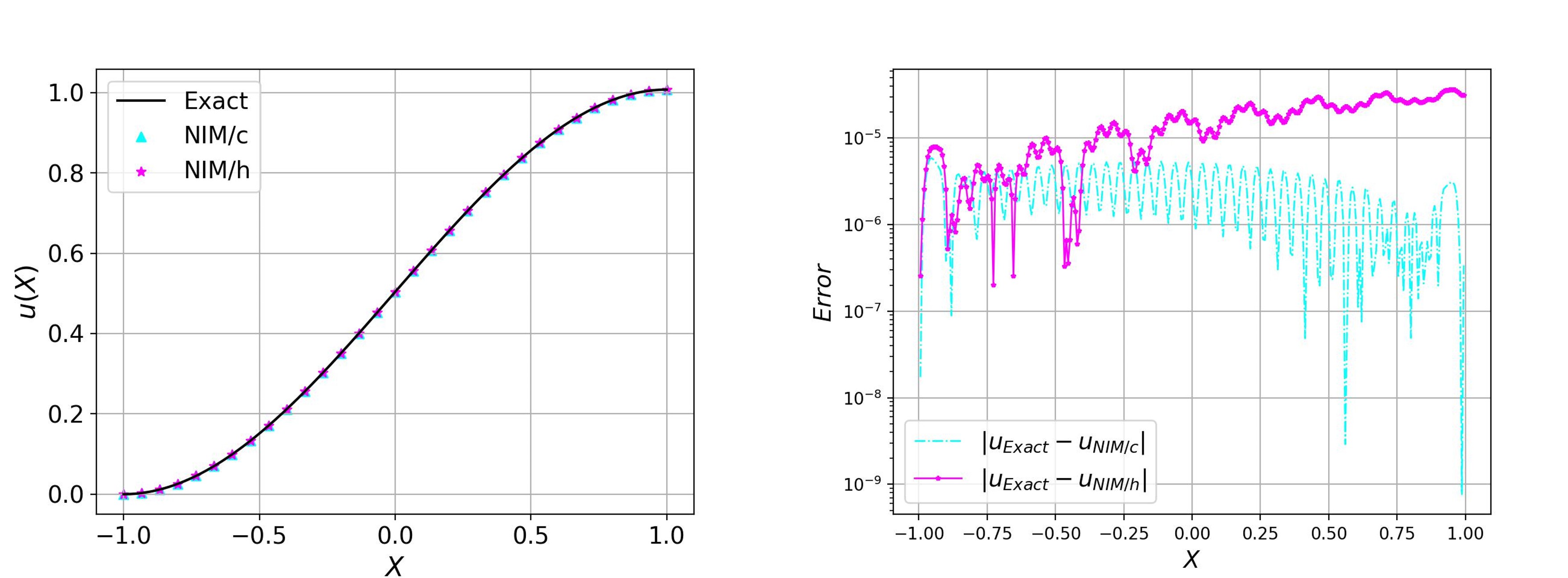}
	\caption{Comparison of the displacement solutions (Left) and the point-wise absolute errors (Right) of NIM/c and NIM/h for the 1D bar problem.}
\label{fig:1dbar_u}
\end{figure}

\begin{figure}[htb]
	\centering
	\includegraphics[angle=0,width=0.8\textwidth]{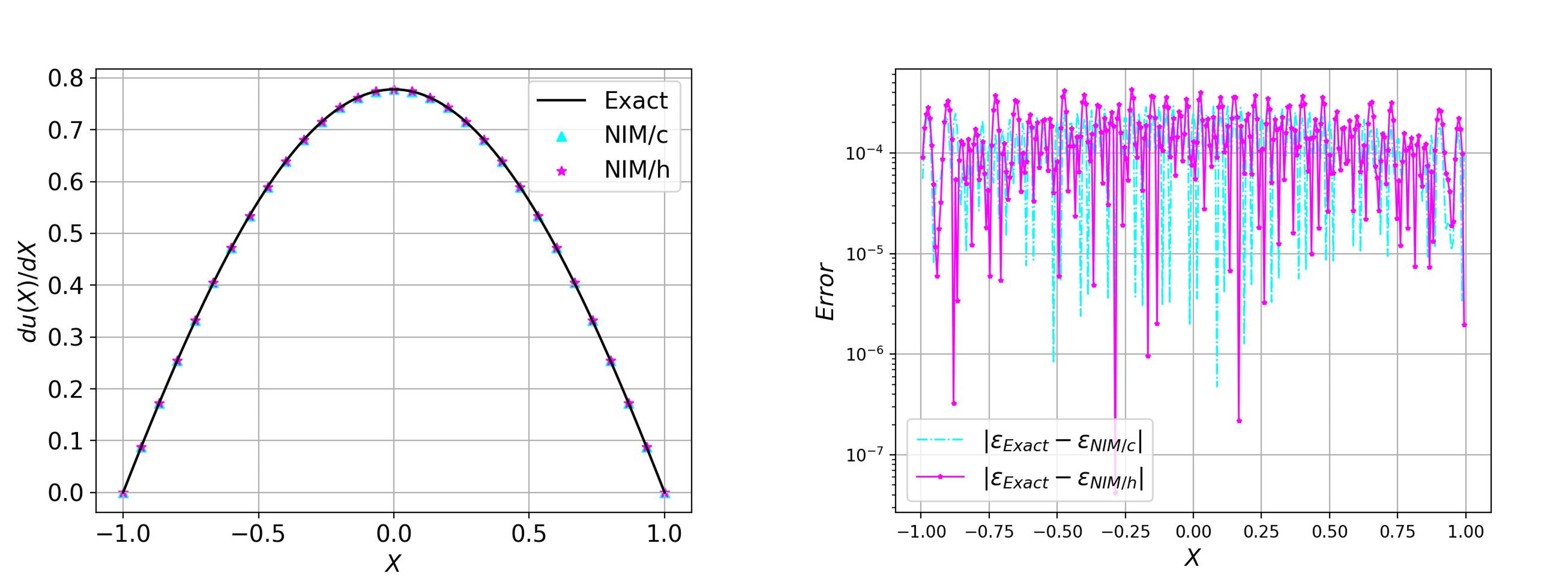}
	\caption{Comparison of the displacement gradient solutions (Left) and the point-wise absolute errors (Right) of NIM/c and NIM/h for the 1D bar problem.}
\label{fig:1dbar_du}
\end{figure}

\begin{figure}[htb]
	\centering
	\includegraphics[angle=0,width=0.8\textwidth]{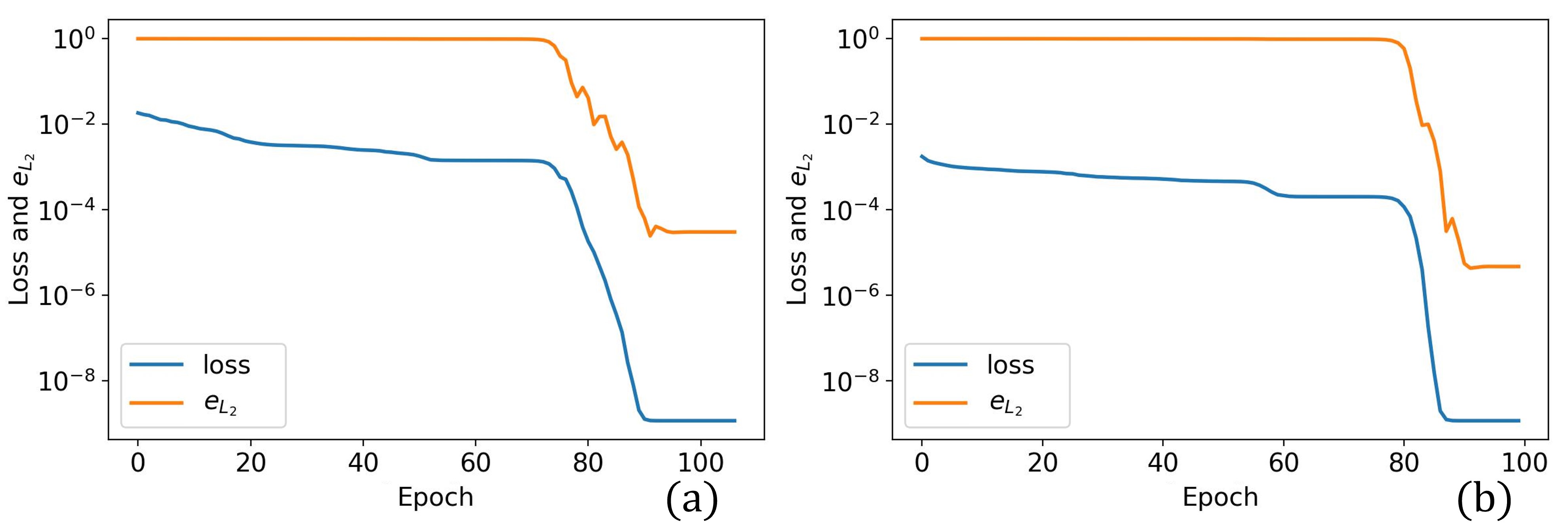}
	\caption{Loss and relative $L_2$ error evolution curves of (a) NIM/h and (b) NIM/c for the 1D bar problem.}
\label{fig:1dbar_tran}
\end{figure}

Figure \ref{fig:1dbar_u} and Figure \ref{fig:1dbar_du} display the point-wise absolute errors in displacement and its gradient of NIM methods, respectively, demonstrating excellent agreement with the analytical solution. 
The comparison of the error plots in Figure \ref{fig:1dbar_u} and \ref{fig:1dbar_du} reveal that NIM/c achieves consistently better accuracy compared to NIM/h, which is attributed to the higher continuity of test functions used in NIM/c. This is consistent with the observation on the linear problems shown in \cite{du2024neural}.
Moreover, the evolution of the loss function and the relative $L_2$ error of these two models against the number of training epochs are given in Figure \ref{fig:1dbar_tran}. It shows that both NIM/c and NIM/h can converge quickly within 90 epochs by using L-BFGS-B training, where the corresponding $e_{L_2}$ are $4.7 \times 10^{-6}$ for NIM/c and $3.0 \times 10^{-5}$ for NIM/h, respectively.
Although only $N_h=41$ nodes are used in the proposed NIM method, it achieves almost the same accuracy as the deep energy method \cite{nguyen2020deep}, which uses $1000$ distributed training points.


\subsection{Uniaxial loading of a 2D plate}\label{sec:plate}
\begin{figure}[htb]
	\centering
	\includegraphics[angle=0,width=0.4\textwidth]{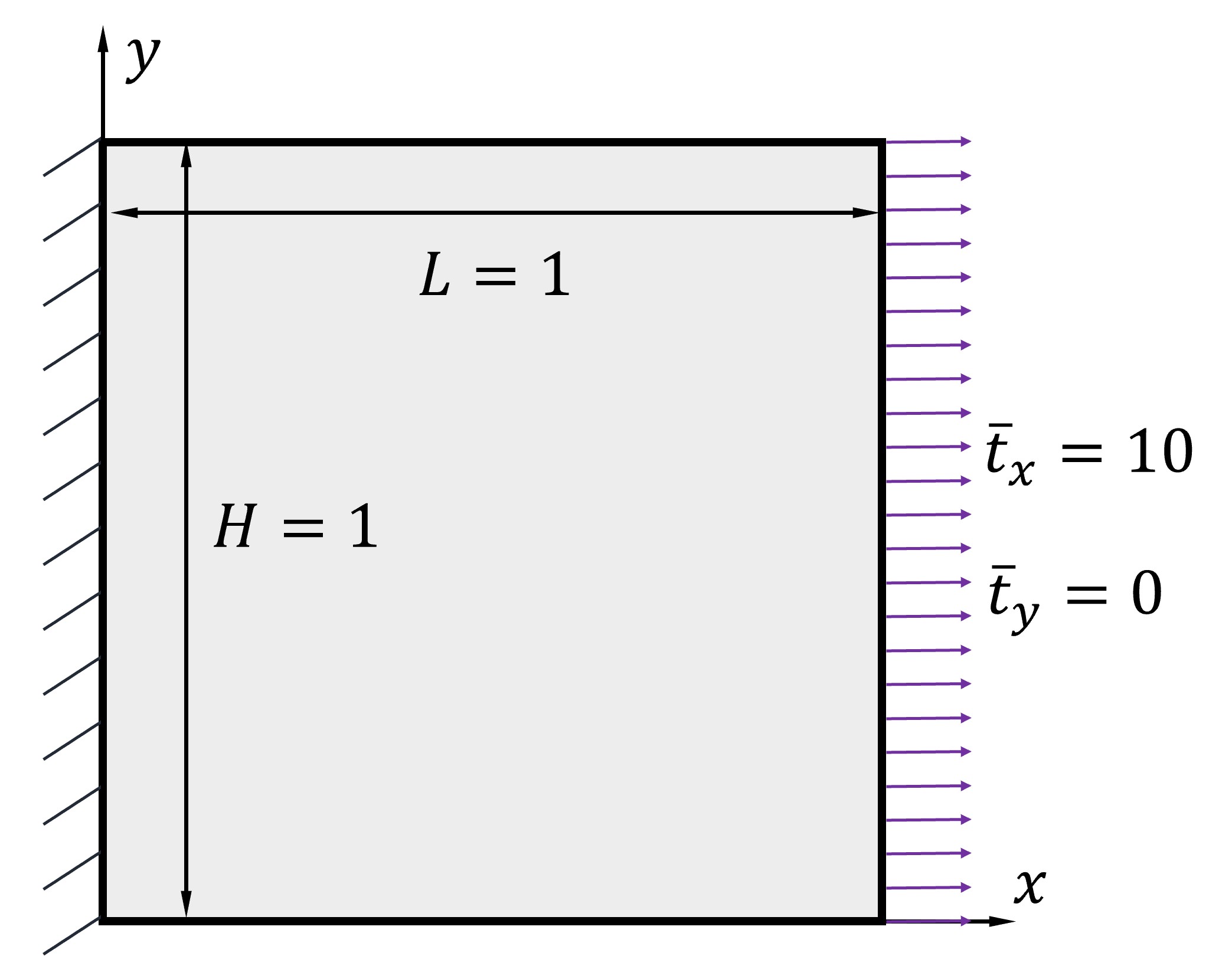}
	\caption{Illustration of the 2D uniaxial loading test.
 }
\label{fig:hyper_2d}
\end{figure}

Next, we consider a simple uniaxial loading 2D problem under plane strain condition, where a square block with dimensions $L=H=1$ m is subjected to a constant load $\bar t_x=10$ N on the right-hand side, with the left-hand side fixed, as depicted in Figure \ref{fig:hyper_2d}.

The material properties are defined with a Young's modulus of $E$ = 1000 $\mathrm{N} / \mathrm{m}^2$, and a Poisson's ratio of $\nu=0.3$, employing a Neo-Hookean material model as detailed in Eqs. \eqref{eq:psi_nh} and \eqref{eq:lame}.
The meshfree discretization of NIM utilizes 441 and 1681 uniformly distributed nodes and subdomains with $\bar r=2.5$.

\begin{figure}[!h]
	\centering
	\includegraphics[angle=0,width=0.75\textwidth]{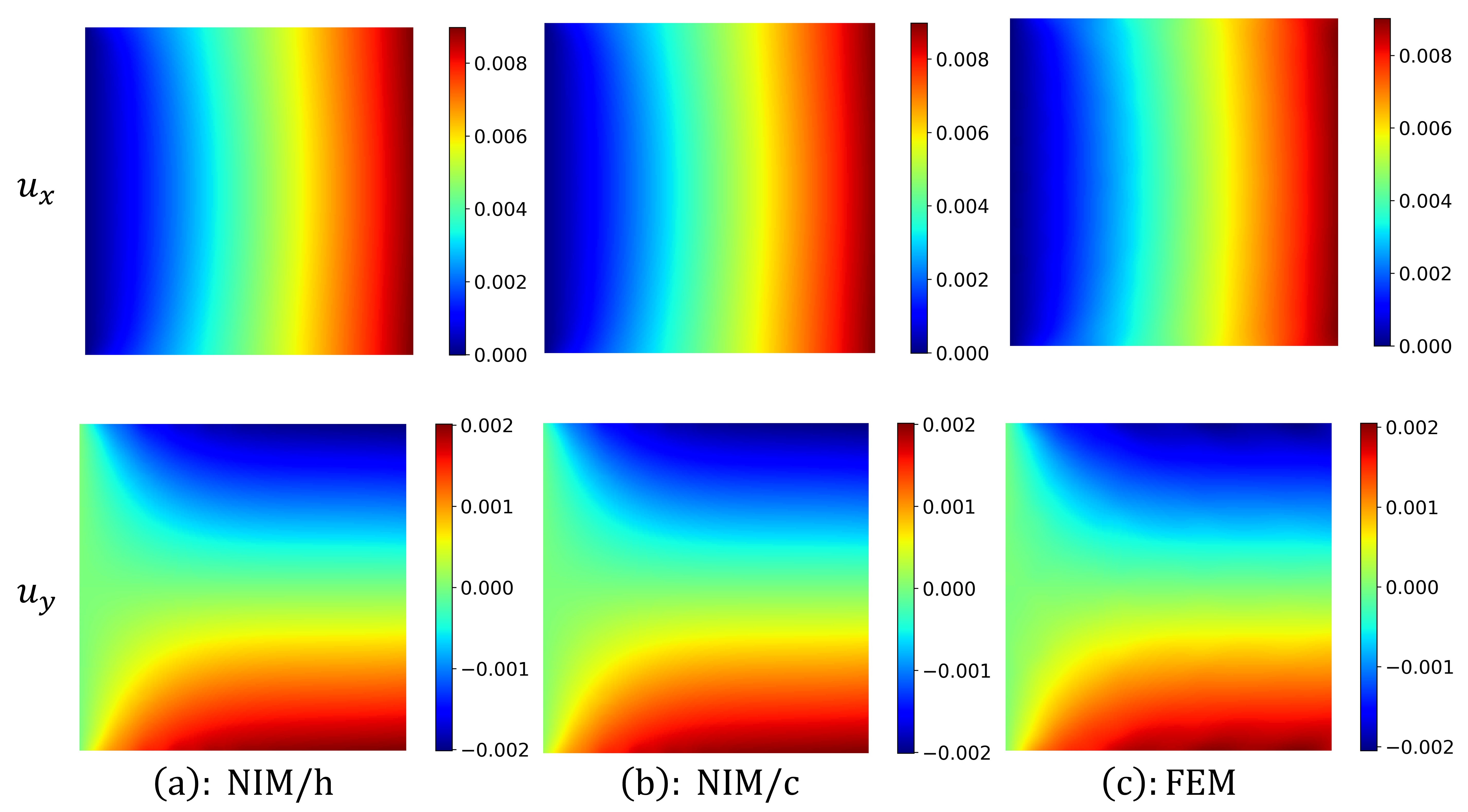}
	\caption{{Displacement of the 2D uniaxial loading test simulated by (a) NIM/h, (b) NIM/c and (c) FEM. }}
\label{fig:hyper_plate_compare}
\end{figure}

\begin{figure}[!h]
	\centering
	\includegraphics[angle=0,width=0.7\textwidth]{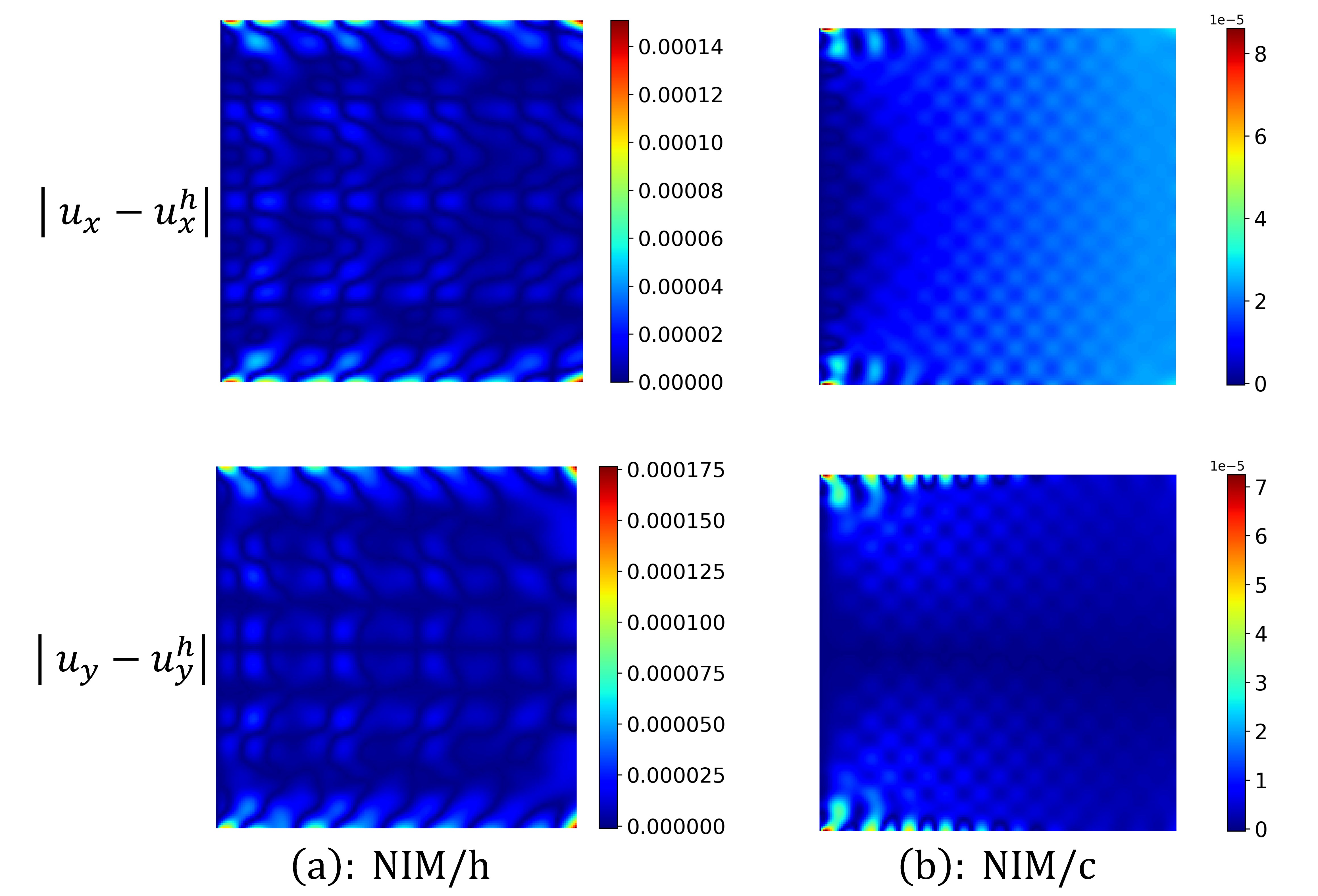}
	\caption{{Comparison of the point-wise absolute errors of the displacement solutions obtained by (a) NIM/h and (b) NIM/c for the 2D uniaxial loading test.}}
\label{fig:hyper_plate_err}
\end{figure}

\begin{figure}[!h]
	\centering
	\includegraphics[angle=0,width=0.8\textwidth]{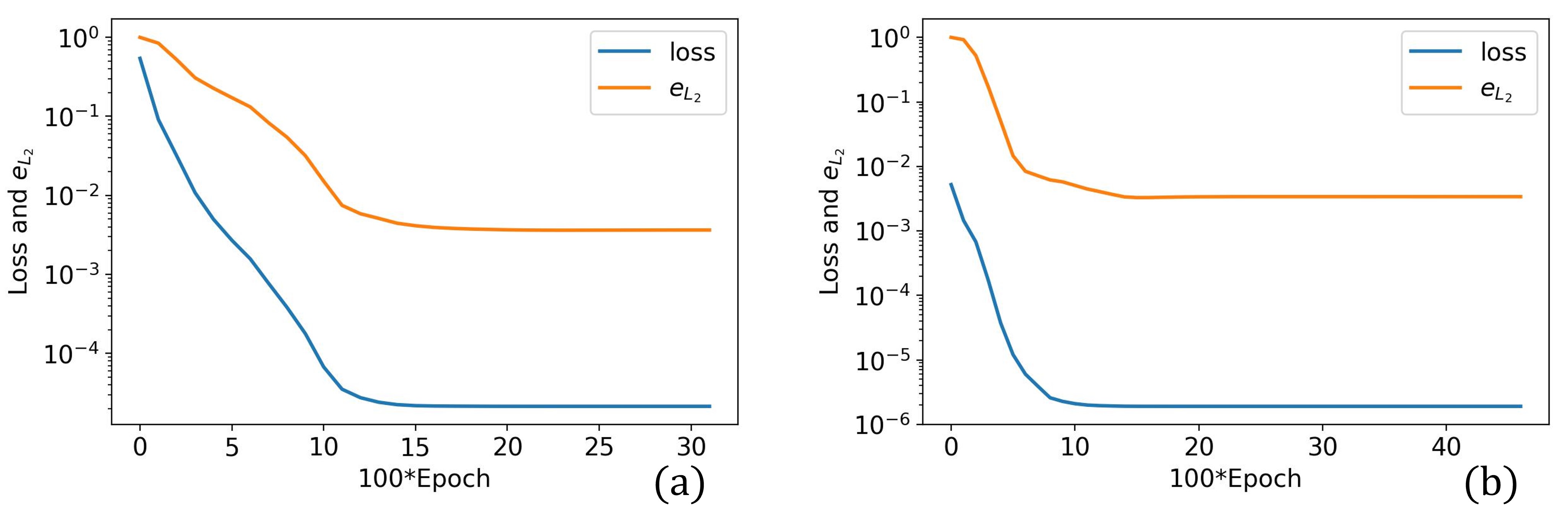}
	\caption{Loss and relative $L_2$ error evolution curves of (a) NIM/h and (b) NIM/c for the 2D uniaxial loading test.}
\label{fig:hyper_plate_loss}
\end{figure}

As displayed in Figure \ref{fig:hyper_plate_compare}, the displacement contours obtained by NIM/h and NIM/c well preserve the symmetries of the solution, and display a smooth transition in displacement without abrupt changes, providing good matches to the FEM reference solution.
Moreover, the point-wise absolute error contours are presented in Figure \ref{fig:hyper_plate_err}. Despite the increased errors near the edges, the overall error magnitudes for both NIM/h and NIM/c remain satisfactory. Specifically, NIM/h achieves an error order of $\mathcal{O}(10^{-4})$, while NIM/c reaches $\mathcal{O}(10^{-5})$. The training history and $e_{L_2}$ evolution, as shown in Figure \ref{fig:hyper_plate_loss}, also confirm that NIM/c achieves slightly higher accuracy than NIM/h, with $e_{L_2}=3.39 \times 10^{-3}$ compared to $e_{L_2}=3.64 \times 10^{-3}$.
This is consistent to the observation in Section \ref{sec:hyper}, NIM/c stands out for its superior accuracy in reproducing complex hyperelastic behaviors, attributed to the higher continuity of the test function. 



\subsection{Large deflection of a cantilever beam subjected to a tip transverse displacement}
\begin{figure}[htb]
	\centering
        \hspace{40pt}
	\includegraphics[angle=0,width=0.6\textwidth]{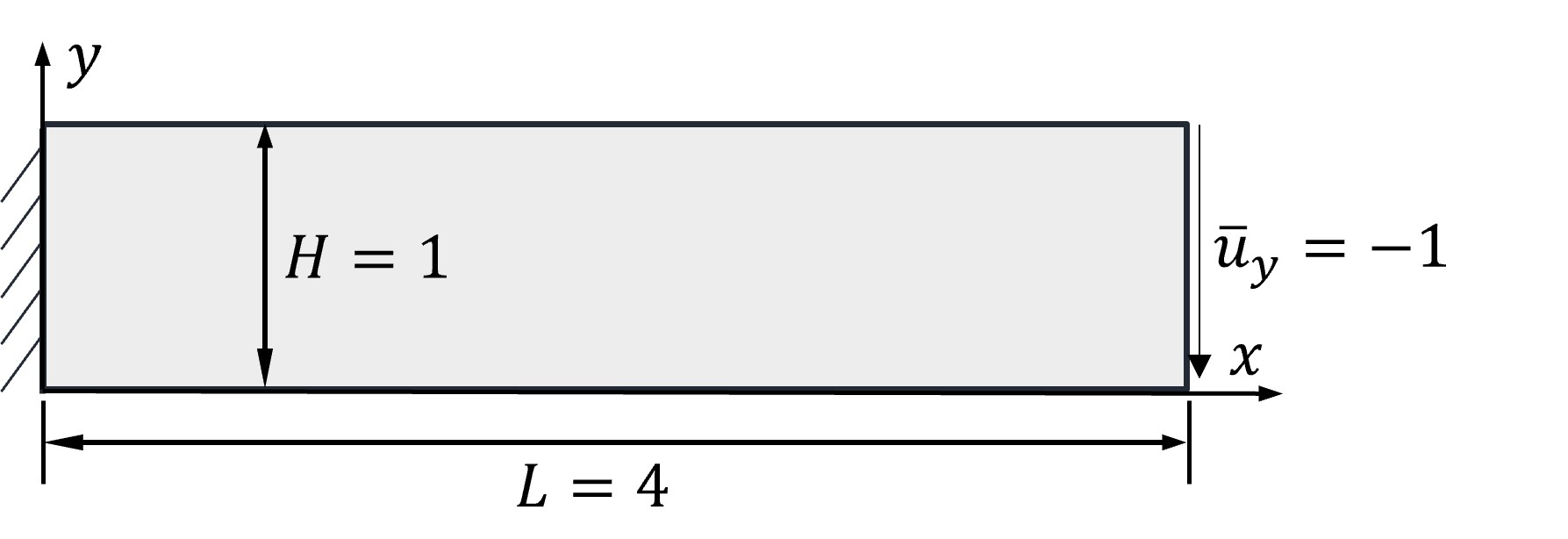}
	\caption{Illustration of the 2D cantilever beam subjected to a tip transverse displacement.}
\label{fig:2dbeam}
\end{figure}

In this example, the large deflection of a cantilever beam under plane strain condition is analyzed, as shown in Figure \ref{fig:2dbeam}, where $L = 4$ m, height $H = 1$ m, and the left side is clamped whereas the right end of the beam is subjected to transverse displacement $\bar u_y=-1$ m. 
Again, the boundary singular kernel method is used in the NeuroPU approximation to enforce these displacement boundary conditions. 

The beam is made of Neo-Hookean hyperelastic material \eqref{eq:psi_nh}, with Young's modulus $E = 1000$ $\mathrm{N} / \mathrm{m}^2$ and Poisson's ratio $\nu=0.3$. 
For NIM modeling, $N_h = 231$ uniformly distributed nodes and $N_{\mathcal{T}} = 861$ subdomains with $\bar r=2.5$ are utilized in this problem.

\begin{figure}[htb]
	\centering
	\includegraphics[angle=0,width=1.0\textwidth]{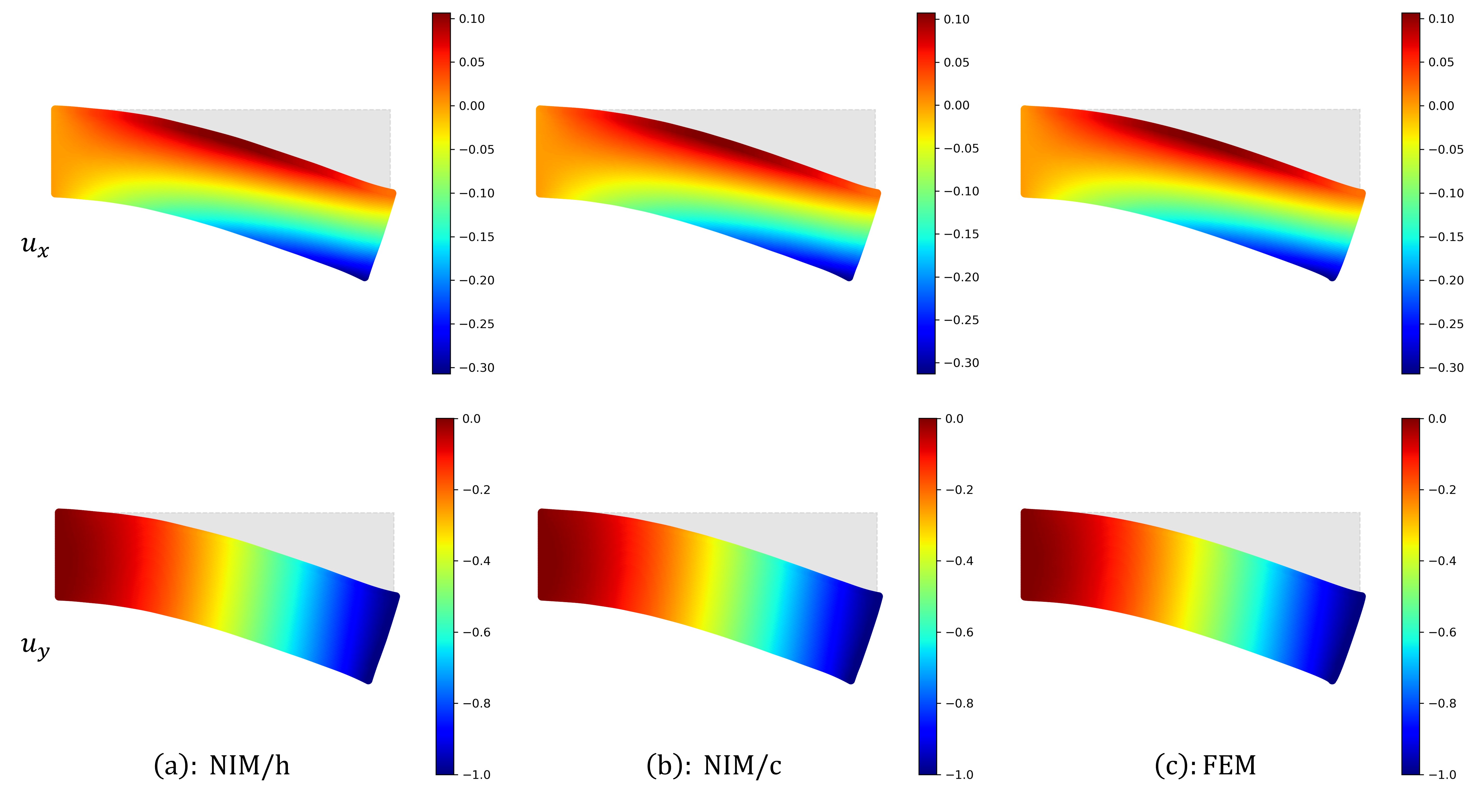}
	\caption{Predicted displacement components ($u_x$ and $u_y$) of the cantilever beam simulated by (a) NIM/h, (b) NIM/c and (c) FEM. Both the undeformed and deformed configurations are illustrated.
    }
\label{fig:2d_beam_deform}
\end{figure}

\begin{figure}[htb]
	\centering
	\includegraphics[angle=0,width=0.9\textwidth]{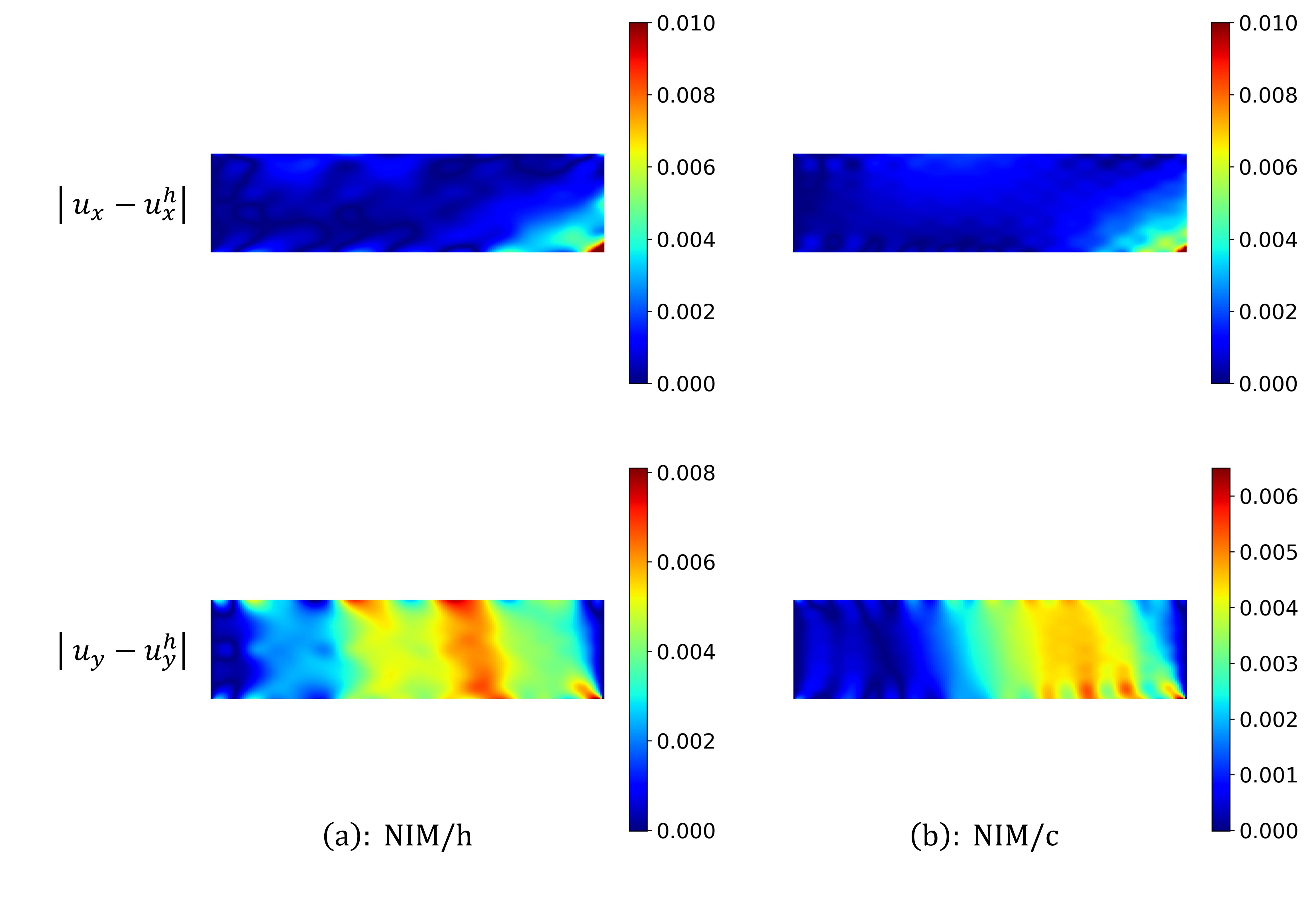}
	\caption{Comparison of the point-wise absolute errors of the displacement solutions obtained by (a) NIM/h and (b) NIM/c for the cantilever beam problem. (Note: The maximum error values of $u_x$ for NIM/h and NIM/c are 0.025 and 0.020, respectively. However, the corresponding colorbar is limited to $[0,0.1]$ for better visualization of the error distribution.)}
\label{fig:2d_beam_err}
\end{figure}

Figure \ref{fig:2d_beam_deform} presents the contour plots of the approximated deformation obtained by NIM/h and NIM/c, in comparison with the FEM reference solution obtained by FEniCS. The close agreements between these results demonstrate the preferable ability of the NIM method to capture the essential mechanical response of hyperelastic materials. 
The point-wise absolute errors of the displacement fields attained by NIM/h and NIM/c are shown in Figure \ref{fig:2d_beam_err}, where the displacement errors of $\mathcal{O}(10^{-2})$ along the x-axis and $\mathcal{O}(10^{-3})$ along the y-axis further corroborate the accuracy of NIM methods.
Additionally, it shows that the essential boundary conditions on the left and right ends are effectively imposed by the boundary singular kernel method, resulting in relatively lower point-wise errors. 
To ensure that the error plot accurately reflects discrepancies across the entire domain, the colorbar in Figure \ref{fig:2d_beam_deform} has been truncated to exclude the extreme error values of $u_x$ at the right-bottom corner, which are 0.025 for NIM/h and 0.020 for NIM/c, respectively. The error in $u_x$ observed in this region may be attributed to the high strain gradient caused by the substantial end-displacement loading, where the differences in discretization resolutions between NIM and FEM would contribute to the solution mismatch in these localized areas.

\begin{figure}[htb]
	\centering
	\includegraphics[angle=0,width=0.9\textwidth]{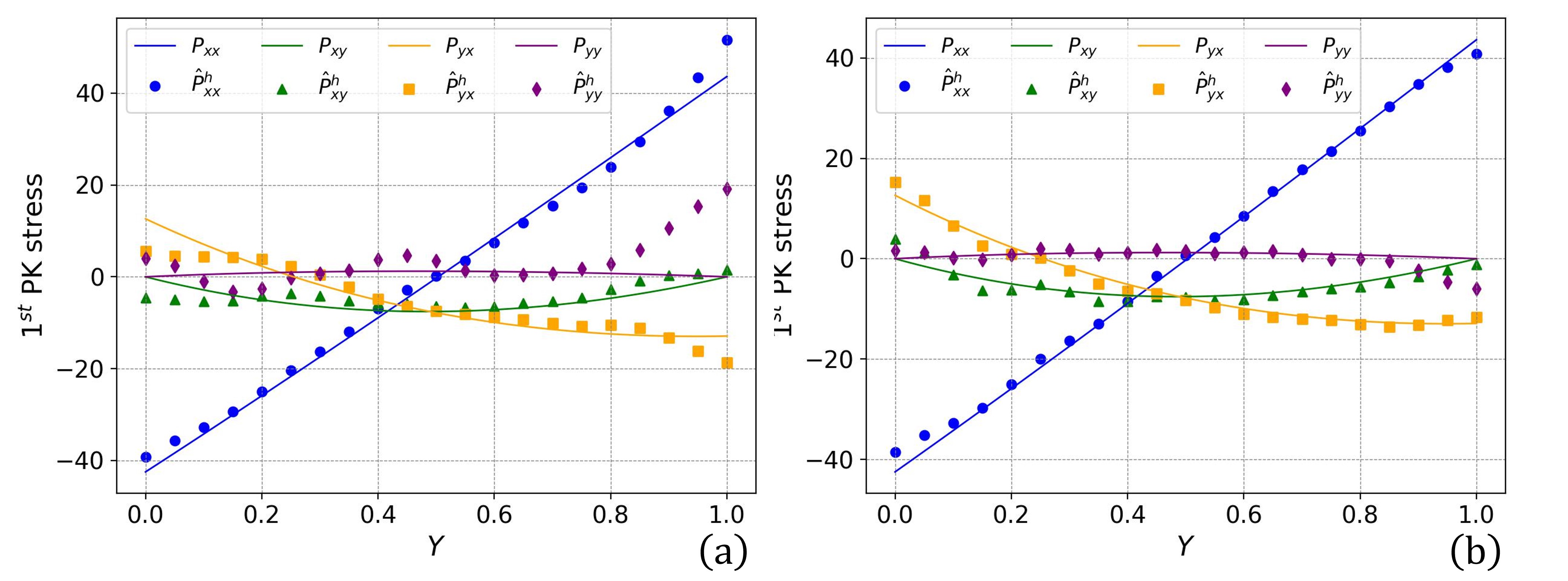}
	\caption{Predicted $1^{st}$ Piola-Kirchhoff stress components ($P_{xx}$, $P_{xy}$, $P_{yx}$, $P_{yy}$) of the cantilever beam along the cross-section on $X=2$ m obtained by (a) NIM/h and (b) NIM/c.}
\label{fig:2d_beam_stress}
\end{figure}

Figure \ref{fig:2d_beam_stress} presents the $1^{st}$ Piola-Kirchhoff stress $\bm{P}$ along the cross-section at $X=2$ m, obtained by NIM/h and NIM/c,  which are compared with the FEM solution. This comparison shows that NIM/h exhibits small oscillations in approximating $P_{yx}$ and $P_{yy}$, especially with notable deviations at the edges ($Y=0$ and $Y=1$). In contrast, NIM/c achieves better alignment with the reference stress values. We believe this is because the higher continuity in test functions helps ensure the traction consistency at domain boundaries. 

\begin{figure}[htb]
	\centering
	\includegraphics[angle=0,width=0.8\textwidth]{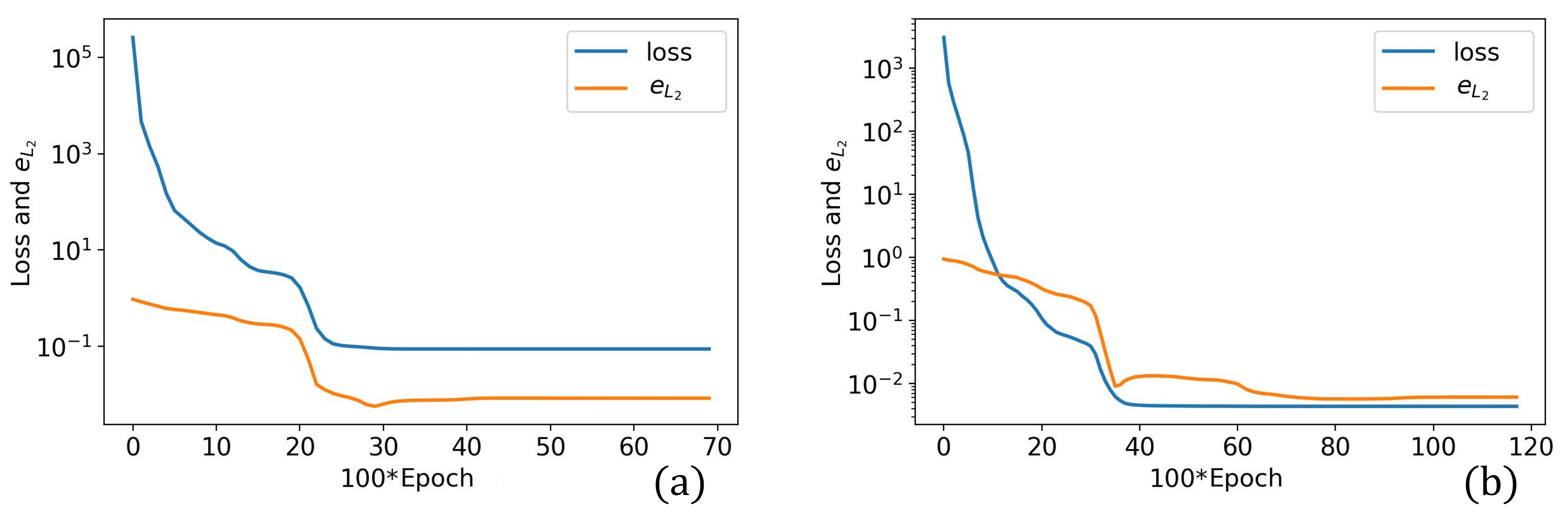}
	\caption{Loss and relative $L_2$ error evolution curves of (a) NIM/h and (b) NIM/c for the cantilever beam problem.}
\label{fig:2d_beam_loss}
\end{figure}

The evolution of the training loss and the relative $L_2$ errors of the NIM solvers are depicted in Figure \ref{fig:2d_beam_loss}, where 
the $e_{L_2}$ error of NIM/h converges to $8.3 \times 10^{-3}$ after 4500 epochs, while for NIM/c, it converges to $6.1 \times 10^{-3}$ after 8000 epochs. 
Overall, in addition to the higher accuracy by NIM/c, consistent with the observations in Sections \ref{sec:hyper} and \ref{sec:plate}, we have also found that it requires more epochs to converge compared to NIM/h. This can be attributed to the distinct properties of the cubic B-spline function and the Heaviside step function. The latter simplifies the integral over subdomains, which affects the accuracy of the local weak form, but also enhances the training efficiency of NIM method.

Overall, the consistently desirable performance in the 2D hyperelasticity problems across different test functions underscores the versatility and effectiveness of the proposed NIM method based on the Petrov-Galerkin framework.

\section{Numerical Results: Inverse Modeling}\label{sec:result_inv}
Accurate estimation of spatially dependent  parameters is crucial for understanding material behaviors, such as the heterogeneous micromechanical properties of biological tissues. 
This is because the reliability of estimated parameters significantly influence the predictive simulation of full-field mechanical responses and structural performance under various loading conditions.
This section is devoted to exploring the NIM method
for inverse modeling of heterogeneous materials,
where the NIM modeling framework in Section \ref{sec:nim}
is extended to the application of identifying heterogeneous material properties through the assimilation of \textit{strain} measurement data.


To verify the effectiveness of NIM for inverse modeling, we examine two synthetic distributions of elastic modulus, as shown in Figure \ref{fig:heter}, which will be further discussed in Section \ref{sec:case1} and \ref{sec:case2}, respectively.
The reference solutions of displacement and strain fields are obtained by using the FEM software FEniCS with the given synthetic material properties.

\subsection{NIM based inverse modeling}
For NIM based inverse modeling setting, in addition to the displacement approximation  \eqref{eq:u_hyper}, an auxiliary NeuroPU model is used to approximate the unknown elastic modulus field $\hat{E}^h(\bm X)$ as described below
\begin{equation}
\hat{E}^h(\boldsymbol{X};\bm \gamma) =  \sum_{I \in \mathcal{S}_x} \Phi_I(\bm{X}) \hat{d}_I^E(\bm{\gamma})
\label{eq:young_pu}
\end{equation}
where $\hat{d}_I^E$ are the nodal coefficients associated with the elastic modulus, and $\Phi_I$ are the shape functions.
It should be emphasized that the shape functions $\Phi_I$ employed in Eq. \eqref{eq:young_pu} may differ from the ones employed for displacement approximation. 
This distinction allows for the selection of different PU functions tailored to specific requirements, such as trigonometric basis functions to describe periodic patterns or Heaviside step functions to handle material discontinuity. As a preliminary investigation, we adopt the same RK shape function (see \ref{sec:RKPM}) to define $\Phi_I(\bm{X})$ for the elastic modulus approximation in this example. 

The modulus nodal coefficients $\hat{d}_I^E(\bm \gamma)$ are obtained from an independent neural network $\mathcal{N}_{\gamma}^E$ parameterized by $\bm{\gamma}$ (see Section \ref{sec:neuro}). To ensure that the values of the approximated  modulus field \eqref{eq:young_pu} remain positive and within a physically realistic range, 
the following sigmoid function is introduced in the output layer of $\mathcal{N}_{\gamma}^E$ to regularize the outputs $\hat{d}_I^{o}$, that is
\begin{equation}
\hat{d}_I^{E}=\frac{3}{1+e^{-\hat{d}_I^{o}}}+0.5
\label{eq:young_sigmoid}
\end{equation}
where $\hat{d}_I^{o}$ is the $I$th output neuron of
$\mathcal{N}_{\gamma}^E$, and therefore,
the range of the nodal coefficients $\hat{d}_I^{E}$ is confined to $[0.5,3.5]$. 
Note that this range and the corresponding constants in \eqref{eq:young_sigmoid} are determined based on the the prior knowledge
of the (synthetic) elastic modulus field. 

In order to assimilate the dataset, one additional loss term is added to the original loss function \eqref{eq:loss_v}
\begin{equation}
\begin{aligned}
\mathcal{L}(\bm \theta, \bm \gamma)&=\frac{1}{N_\mathcal{T}}\sum_{k=1}^{N_v}\sum_{s=1}^{N_\mathcal{T}}\left\|{\mathcal{R}}_s^{(k)} \right\|^2 +  \frac{\alpha}{N_\text{data}}\sum_{j=1}^{N_\text{data}}\left\|{\bm F(\bm X_j)- \hat {\bm F}^h(\bm X_j)}\right\|^2
\label{eq:loss_data}
\end{aligned}
\end{equation}
where $\bm F (\bm X_j) = \{F{xx} (\bm X_j),F{xy} (\bm X_j),F{yx} (\bm X_j),F{yy} (\bm X_j) \}$ represents the measurements of deformation gradient tensor on $\bm X_j$, $\alpha$ is the penalty number to balance the terms associated with the data loss and the physical residual loss. $N_\text{data}$ is the number of data points selected from the strain field.
Here, for convenience, the deformation gradient data is also referred to as strain data, with a slight abuse of terminology.

\subsection{Case study} \label{sec:case_inverse}

In this section, a unit square specimen made of St. Venant-Kirchhoff hyperelastic material as described in Eq. \eqref{eq:psi_st} is considered.
The Poisson's ratio is specified as 0.35, while the heterogeneous Young's modulus field in $\Omega = [0,1] \times [0,1]$ is represented by the two synthesis cases as depicted in Figure \ref{fig:heter}.
The boundary conditions are set as equibiaxial displacement of $0.2$ on all four sides.

\subsubsection{Case 1: Symmetric elastic modulus} \label{sec:case1}

\begin{figure}[htb]
	\centering
	\includegraphics[angle=0,width=0.8\textwidth]{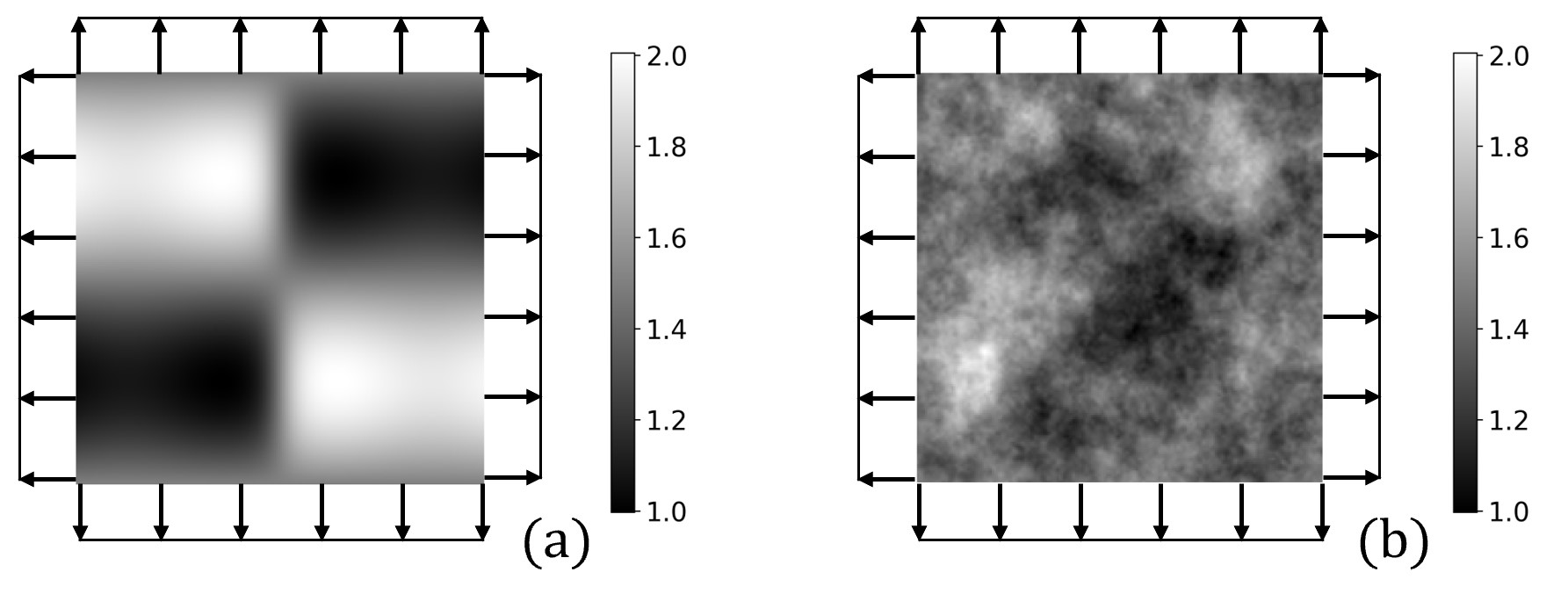}
	\caption{Two exemplary elastic modulus fields defined in $\Omega = [0,1] \times [0,1]$: (a) Symmetric elastic modulus; (b) Elastic modulus represented by Gaussian
random field distributions.}
\label{fig:heter}
\end{figure}

In this case, the following equation is employed to define the symmetric Young's modulus field
    
\begin{equation}
    E^{\text {exact }}(X, Y)=[0.1 \sin (2 \pi X)+\tanh (10 X)] \times \sin (2 \pi Y)
    \label{eq:analytical_young}
    \end{equation}
Note that the ground truth $E^{\text {exact }}(X, Y)$ will be normalized to fall within the range [1, 2], and the normalized distribution is depicted in Figure \ref{fig:heter}a.

In the NIM inverse modeling, a uniform discretization of $N_h = 961$ nodes are used for both approximation of the elastic modulus field $\hat{E}^h$ in \eqref{eq:young_pu} and the displacement field in \eqref{eq:u_hyper}.  
$N_{\mathcal{T}} = 2601$ uniformly distributed subdomains with $\bar r=2.0$ are employed to construct the loss function \eqref{eq:loss_data}.
We randomly select $N_\text{data}=3000$ data points from the ground truth strain fields and set the penalty number to $\alpha=10$ in Eq. \eqref{eq:loss_data} to train the NIM model, whereby both the displacement solution and elastic modulus estimation are obtained.

\begin{figure}[htb]
	\centering
	\includegraphics[angle=0,width=1.0\textwidth]{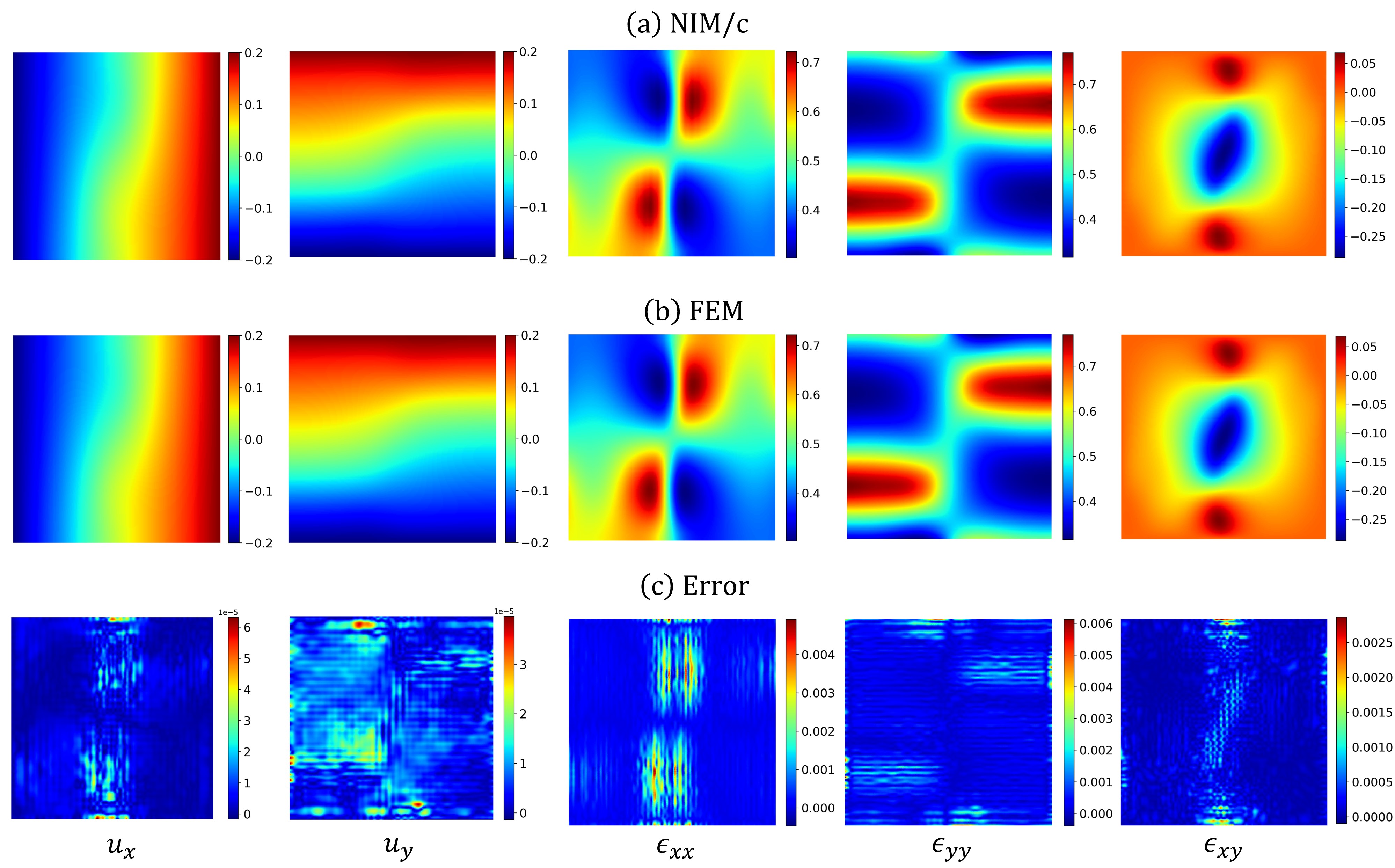}
	\caption{Comparison of displacement ($u_x$ and $u_y$) and Green-Lagrangian strain ($\epsilon_{xx}$, $\epsilon_{yy}$ and $\epsilon_{xy}$) fields obtained from the NIM and FEM solvers for the case with symmetric elastic modulus field.}
\label{fig:case1_err_dis}
\end{figure}

\begin{figure}[htb]
	\centering
	\includegraphics[angle=0,width=0.8\textwidth]{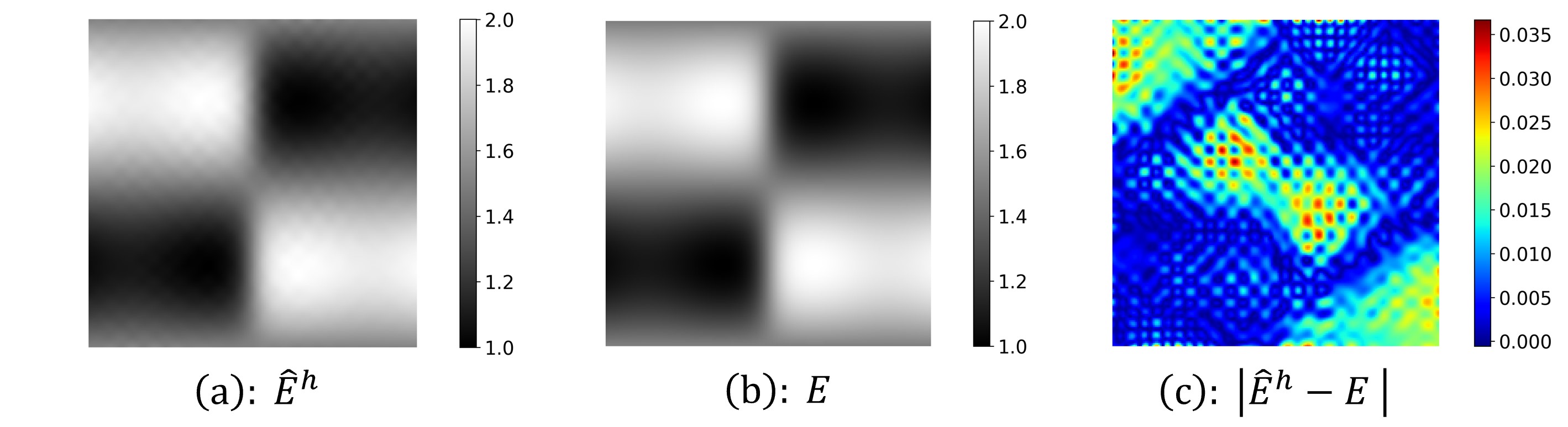}
	\caption{Distributions of (a) the estimated elastic modulus field obtained from the NIM solver, (b) the ground truth for the symmetric elastic modulus, and (c) the point-wise absolute error.}
\label{fig:case1_err_young}
\end{figure}

 The predicted solutions and the parameter estimation by NIM/c are presented in Figure \ref{fig:case1_err_dis} and \ref{fig:case1_err_young}, respectively.
 In Figure \ref{fig:case1_err_dis}, the  predicted displacement ($u_x$ and $u_y$ shown in the first two columns) and Green-Lagrangian strain\footnote{With a slight abuse of notation, the symbol 
$\bm \epsilon$ is used to denote the Green-Lagrangian strain tensor instead of the commonly used $\bm E$, to avoid confusion with elastic modulus $E$.} 
 fields ($\epsilon_{xx}$, $\epsilon_{yy}$ and $\epsilon_{xy}$ shown in the last three columns) are compared against the FEM reference solutions,  
 where the distributions of the point-wise absolute errors are provided in Figure \ref{fig:case1_err_dis}c.
 It is observed that NIM provides highly accurate predictions of both the displacement and strain distributions, with absolute errors at the level of $\mathcal{O}(10^{-5})$ for displacements and $\mathcal{O}(10^{-3})$ for strains. We note that although deformation gradient data are incorporated into the training, the overall strain prediction accuracy remains lower than that of displacement. This discrepancy is attributed to the differentiation operation, which reduces approximation accuracy, and to the more complex strain fields, which complicates the approximation process.




 Moreover, the estimated elastic modulus field $\hat E^h$ obtained by the NIM solver is depicted in Figure \ref{fig:case1_err_young}, where the ground truth $E$ for the symmetric elastic modulus and the corresponding absolute error are also provided for comparison.
 As shown in Figure \ref{fig:case1_err_young}c, the maximum absolute error $\left|\hat E^h-E\right|$ is smaller than 0.04, corresponding to approximately $2 \%$ relative error.
 This close agreement demonstrates that the NIM inverse modeling scheme can achieve accurate parameter estimation using only strain data,
 without the need for direct modulus measurements.
It is also observed that relatively large errors appear in the corners and central regions where there are sharp gradients in modulus values. This suggests that more data or refined discretization may be needed in these areas to enhance the accuracy of the estimation results.
 
\begin{figure}[htb]
	\centering
	\includegraphics[angle=0,width=0.85\textwidth]{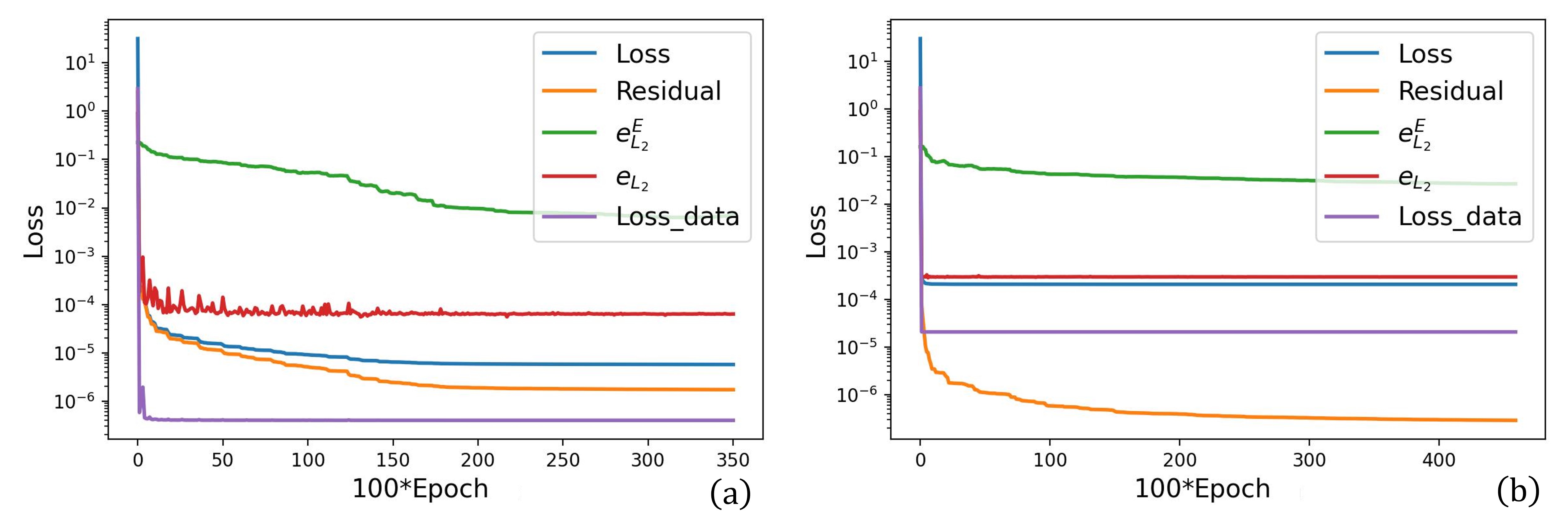}
	\caption{Training trajectories of loss functions (including the total loss, and the individual components corresponding to residuals and measurement data) and the relative $L_2$ error of estimated displacement ($e_{L_2}$) and modulus ($e_{L_2}^E$) obtained by NIM/c inverse modeling for two cases: (a) symmetric elastic modulus, and (b) elastic modulus represented by Gaussian random field (GRF).}
\label{fig:inverse_loss}
\end{figure}

 The training trajectories of losses and the associated relative $L_2$ errors are provided in Figure \ref{fig:inverse_loss}a. 
 It shows that the relative $L_2$ error of displacement, $e_{L_2}$, decreases to $6.31 \times 10^{-5}$ after 10,000 epochs,
while the relative $L_2$ error of elastic modulus, $e^E_{L_2}$, reaches the converged value $6.62 \times 10^{-3}$ after 30,000 epochs. 
These results emphasize the applicability of the NIM method for estimating the hidden mechanical properties by assimilating datasets that are not directly relevant to the unknown fields.

\subsubsection{Case 2: Gaussian random field} \label{sec:case2}
In this case, 
we generate a synthetic distribution of elastic modulus using a 
2D Gaussian random field (GRF) generator\footnote{https://github.com/bsciolla/gaussian-random-fields}, where $\alpha =3$ is selected as the power of the power-law momentum distribution,
as illustrated in Figure \ref{fig:heter}b. 
This field mimics the complex heterogeneous characteristics of biological tissues \cite{Wu2024-bg,HE2024106443}.
For NIM-based inverse modeling, the distribution of nodes and subdomains as well as the penalty coefficient used in the loss function remain consistent with those described in Section \ref{sec:case1}. 
Nevertheless, to better capture the increased complexities of the elastic modulus field and the resulting strain fields,
we have chosen to use more data points, i.e., $N_\text{data}=5000$, for model training.

\begin{figure}[htb]
	\centering
	\includegraphics[angle=0,width=1.0\textwidth]{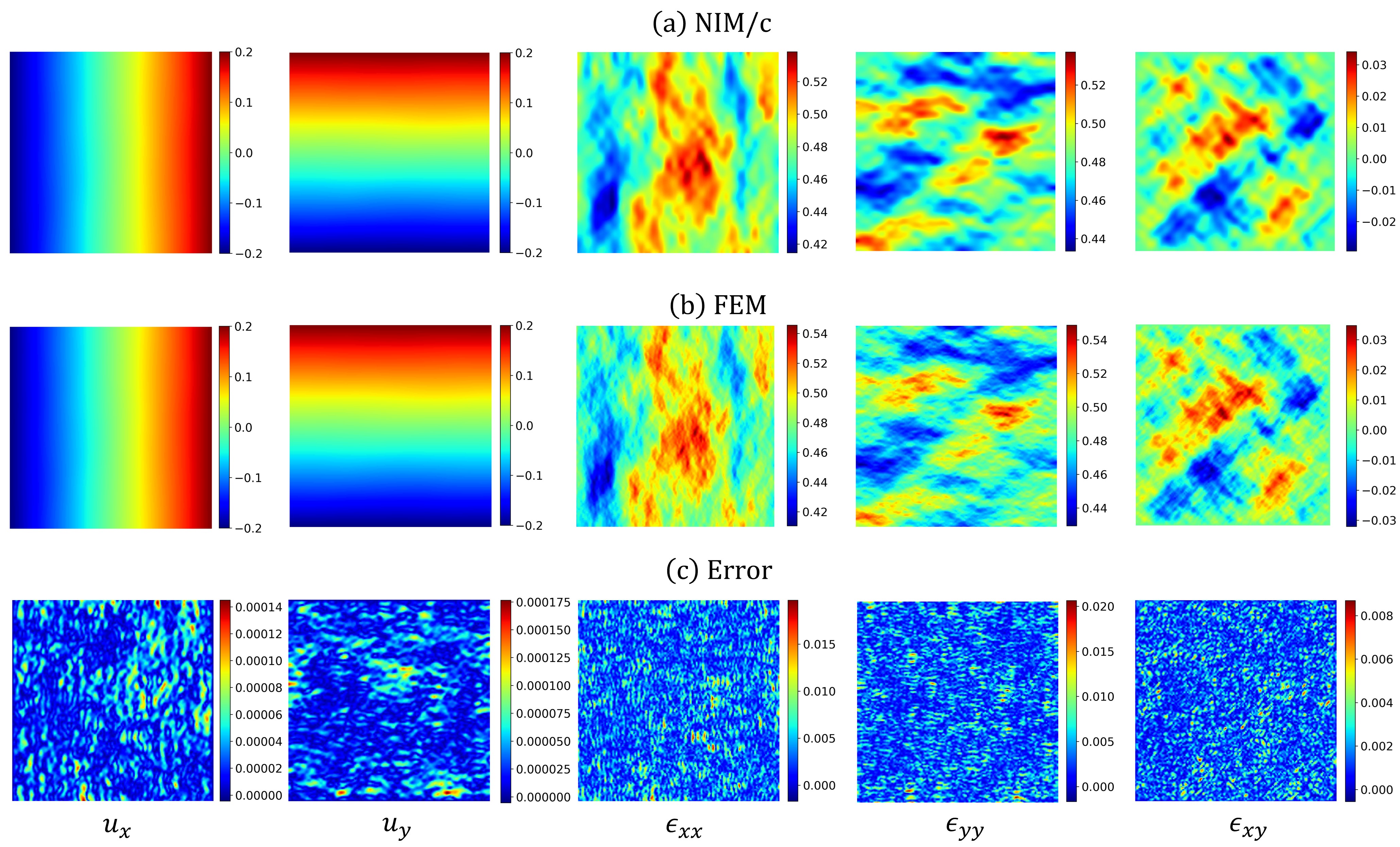}
	\caption{Comparison of displacement ($u_x$ and $u_y$) and Green-Lagrangian strain ($\epsilon_{xx}$, $\epsilon_{yy}$ and $\epsilon_{xy}$) fields obtained from the NIM and FEM solvers for the case with elastic modulus field represented by GRF.}
\label{fig:case2_err_dis}
\end{figure}

\begin{figure}[htb]
	\centering
	\includegraphics[angle=0,width=0.8\textwidth]{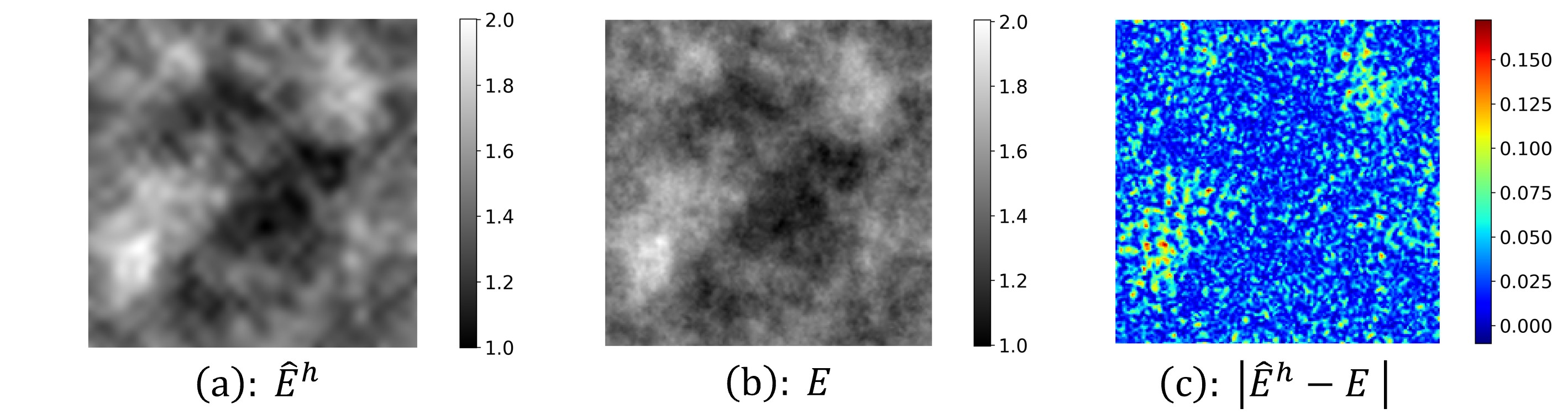}
	\caption{Distributions of (a) the estimated elastic modulus field obtained from the NIM solver, (b) the ground truth for elastic modulus field represented by GRF, and (c) the point-wise absolute error.}
\label{fig:case2_err_young}
\end{figure}

The approximated displacement and strain fields obtained by NIM are compared with the FEM reference solutions in Figure \ref{fig:case2_err_dis}.
It can be seen that the maximum absolute errors for displacement and strain are as small as $\mathcal{O}(10^{-4})$ and $\mathcal{O}(10^{-2})$, respectively. 
We observe that the error levels in this case are relatively higher compared to those in Case 1 (see Section \ref{sec:case1}). This increase can be attributed to the higher complexity of the elastic modulus distribution. 
Additionally, the estimated elastic modulus field is shown in Figure \ref{fig:case2_err_young}. Although the underlying modulus with random features
is difficult to capture, 
the proposed NIM scheme still manage to achieve satisfactory estimation with a maximum absolute error less than 0.16 (i.e., $8 \%$ in relative error). This demonstrate the excellent efficacy of NIM for inverse modeling even without direct measurements.


The training history, presented in Figure \ref{fig:inverse_loss}b, demonstrates that, both the estimated displacement and elastic modulus fields converge steadily even with an unknown complex parameter field. The displacement solution reaches its convergent value of the relative $L_2$ error after approximately 1,000 epochs ($e_{L_2}=2.96 \times 10^{-4}$), while the elastic modulus field converges after 43,000 epochs ($e^E_{L_2}=2.72 \times 10^{-2}$).
The comparison of training histories reveals that the learning of displacement converges faster than parameter estimation in the data assimilation setting. This disparity in convergence rates, consistent with the findings from Case 1 (Figure \ref{fig:inverse_loss}a), can be primarily attributed to the training relying solely on strain data, with no modulus measurements available.

\begin{figure}[htb]
	\centering
	\includegraphics[angle=0,width=0.8\textwidth]{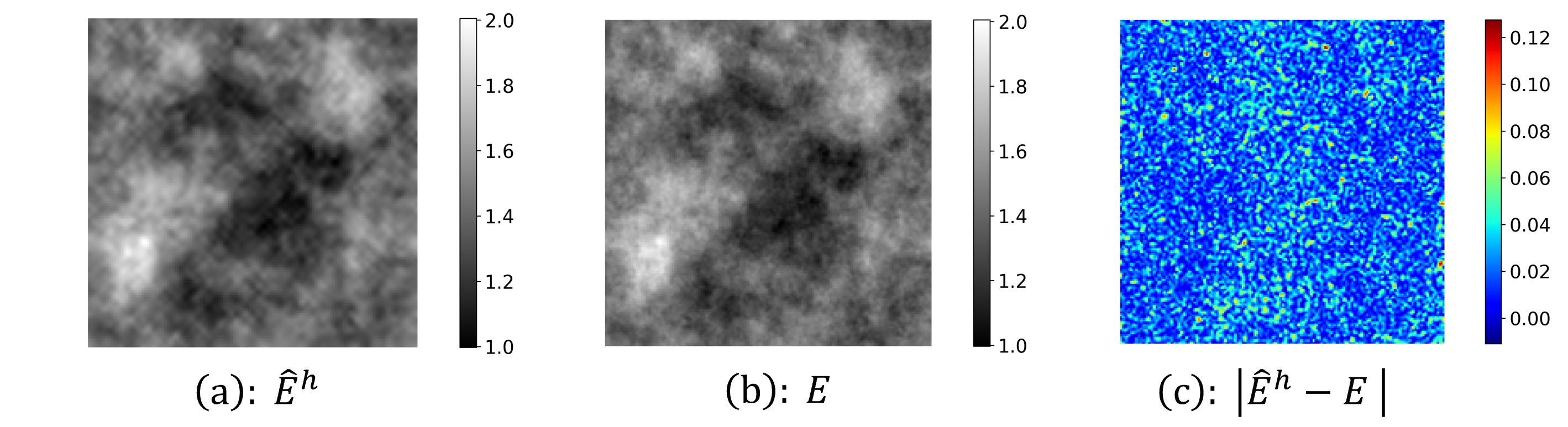}
	\caption{Distributions of (a) the estimated elastic modulus field obtained from the NIM solver with 3721 nodes and 6561 subdomains, (b) the ground truth for elastic modulus field represented by GRF, and (c) the point-wise absolute error.}
\label{fig:case2_err_young_more}
\end{figure}

The performance of NIM-based inverse modeling can be further improved with increasing the underlying discretization resolution. 
To demonstrate that, we also present the results obtained with 3721 nodes and 6561 subdomains in Figure \ref{fig:case2_err_young_more},
which shows a more expressive  approximation space for the elastic modulus field yields a significant error reduction in parameter identification, compared to Figure \ref{fig:case2_err_young}.
Consequently, we note that more complex Young's modulus fields usually require a larger number of nodes in the NeuroPU approximation for accurate representation.


\FloatBarrier
\section{Conclusion}\label{sec:conclusion}
In this study, we extend NIM, a differentiable programming-based AI methodology, to forward and inverse modeling of finite-strain problems. We integrate hyperelasticity models into NIM's loss function using local variational formulations and implement it with the JAX framework.
A hybrid neural-numerical approach coupling DNNs with partition-of-unity basis functions, NeuroPU, is adopted for solution approximation to improve the accuracy and training efficiency.
The introduction of the boundary singular kernel method into NeuroPU enables direct enforcement of essential boundary conditions without additional parameters or penalty terms. 

NIM offers several advantages over other differentiable programming-based methods, notably preserving meshfree properties by constructing the loss function across overlapping subdomains using a Petrov-Galerkin approach, which eliminates the need for conforming meshes in domain integration. Additionally, the loss function, based on the consistent local variational formulation, allows easy adaptation to various hyperelastic material models without structural modifications. Furthermore, NIM bypasses the need for consistent material tangent stiffness derivations, simplifying the simulation process.


Our numerical experiments demonstrate NIM's effectiveness in both forward and inverse modeling of hyperelastic materials. In forward modeling, NIM accurately predicts nonlinear elastic deformations for Neo-Hookean and St. Venant-Kirchhoff hyperelasticity with relatively few training epochs, even without labeled data, highlighting its capability in efficiently simulating complex nonlinear materials. For inverse modeling, NIM successfully identifies hidden material properties in heterogeneous hyperelastic materials using only strain field data for assimilation. This process requires just 30k-40k training epochs with the L-BFGS-B optimizer, thanks to the improved efficiency afforded by using NeuroPU to approximate the unknown material property distributions, which significantly streamlines the training process. Furthermore, we show that improved identification accuracy can be achieved by refining NeuroPU discretizations for the elastic modulus field. 

In conclusion, these results underscore NIM's versatility and efficiency in handling both forward prediction and inverse parameter identification in nonlinear elasticity problems.
NIM serves as a promising tool for predicting the mechanical response and studying material properties within a flexible, consistent, and differentiable framework.
\appendix
\section{Reproducing Kernel Shape Function}\label{sec:RKPM}
The reproducing kernel (RK) shape function $\Psi_I(\mathbf{x})$ is defined as
\begin{equation}\label{eq:RK_shape}
\Psi_I(\boldsymbol x)=\boldsymbol p^{[p] T}\left(\boldsymbol x_I-\boldsymbol x\right) \bm{b}(\boldsymbol x) \phi_a\left(\boldsymbol x_I-\boldsymbol x\right)
\end{equation}
where $\phi_a$ represents the kernel function that ensures the smoothness of the RK approximation function and the compactness with support size $a$. In this study, the cubic B-spline function that preserves $C^2$ continuity is selected to be the kernel function

\begin{equation}\label{eq:kernel_B}
\phi_a(z)= \begin{cases}\frac{2}{3}-4 z^2+4 z^3 & 0 \le z \leq \frac{1}{2} \\ 
\frac{4}{3}-4 z+4 z^2-\frac{4}{3} z^3 & \frac{1}{2}< z\leq 1 \\
0 & z > 1 
\end{cases}
\end{equation}
with $z=\lVert \boldsymbol x_I-\boldsymbol x \rVert / a$.
In Eq. \eqref{eq:RK_shape}, $\boldsymbol{p}^{[p]}(\boldsymbol{x})$ is a vector of monomial basis functions up to the $p$th order
\begin{equation}\label{eq:basis}
\boldsymbol p^{[p]}(\boldsymbol x)=\left\{1, x_1, x_2, x_3, \cdots, x_1^i x_2^j x_3^k, \ldots, x_3^p\right\}^T, 0 \leq i+j+k \leq p
\end{equation}

Then we substitute the $p$th order reproducing conditions, 
\begin{equation}
\sum_{I \in \mathcal{S}_x} \Psi_I(\boldsymbol{x}) \boldsymbol{p}^{[p]}\left(\boldsymbol x_I\right)=\boldsymbol{p}^{[p]}(\boldsymbol{x})
\label{eq:repro}
\end{equation}
to Eq. \eqref{eq:RK_shape} to determine the unknown parameter vector $\boldsymbol{b}(\boldsymbol{x})$, which yields

\begin{equation}
\bm{b}(\boldsymbol x)=\boldsymbol A^{-1}(\boldsymbol{x}) \boldsymbol{p}^{[p]}(\boldsymbol{0})
\label{eq:C}
\end{equation}
where $\bm{A}(\bm{x})$ is the moment matrix
\begin{equation}
\bm A(\boldsymbol x)=\sum_{I \in \mathcal{S}_x} \boldsymbol p^{[p]}\left(\boldsymbol x_I-\boldsymbol x\right) \boldsymbol p^{[p] T}\left(\boldsymbol x_I-\boldsymbol x\right) \phi_a \left(\boldsymbol x_I-\boldsymbol x\right)
\label{eq:A}
\end{equation}
Invoking Eq. \eqref{eq:C} into Eq. \eqref{eq:RK_shape}, we have the following expression for the RK shape functions:
\begin{equation}
\Psi_I(\boldsymbol x)=\boldsymbol p^{[p] T}(\boldsymbol x) \boldsymbol A^{-1}(\boldsymbol x) \boldsymbol p^{[p]}\left(\boldsymbol x_I-\boldsymbol x\right) \phi_a\left(\boldsymbol x_I-\boldsymbol x\right)
\end{equation}

\newpage

\bibliographystyle{cas-model2-names} 

\bibliography{ref_diff,ref_Meshfree,ref_new}

\begin{thebibliography}{76}
\expandafter\ifx\csname natexlab\endcsname\relax\def\natexlab#1{#1}\fi
\providecommand{\url}[1]{\texttt{#1}}
\providecommand{\href}[2]{#2}
\providecommand{\path}[1]{#1}
\providecommand{\DOIprefix}{doi:}
\providecommand{\ArXivprefix}{arXiv:}
\providecommand{\URLprefix}{URL: }
\providecommand{\Pubmedprefix}{pmid:}
\providecommand{\doi}[1]{\href{http://dx.doi.org/#1}{\path{#1}}}
\providecommand{\Pubmed}[1]{\href{pmid:#1}{\path{#1}}}
\providecommand{\bibinfo}[2]{#2}
\ifx\xfnm\relax \def\xfnm[#1]{\unskip,\space#1}\fi
\bibitem[{Abadi et~al.(2016)Abadi, Barham, Chen, Chen, Davis, Dean, Devin, Ghemawat, Irving, Isard et~al.}]{abadi2016tensorflow}
\bibinfo{author}{Abadi, M.}, \bibinfo{author}{Barham, P.}, \bibinfo{author}{Chen, J.}, \bibinfo{author}{Chen, Z.}, \bibinfo{author}{Davis, A.}, \bibinfo{author}{Dean, J.}, \bibinfo{author}{Devin, M.}, \bibinfo{author}{Ghemawat, S.}, \bibinfo{author}{Irving, G.}, \bibinfo{author}{Isard, M.}, et~al., \bibinfo{year}{2016}.
\newblock \bibinfo{title}{Tensorflow: a system for large-scale machine learning.}, in: \bibinfo{booktitle}{Osdi}, \bibinfo{organization}{Savannah, GA, USA}. pp. \bibinfo{pages}{265--283}.
\bibitem[{Abueidda et~al.(2022)Abueidda, Koric, Al-Rub, Parrott, James and Sobh}]{abueidda2022deep}
\bibinfo{author}{Abueidda, D.W.}, \bibinfo{author}{Koric, S.}, \bibinfo{author}{Al-Rub, R.A.}, \bibinfo{author}{Parrott, C.M.}, \bibinfo{author}{James, K.A.}, \bibinfo{author}{Sobh, N.A.}, \bibinfo{year}{2022}.
\newblock \bibinfo{title}{A deep learning energy method for hyperelasticity and viscoelasticity}.
\newblock \bibinfo{journal}{European Journal of Mechanics-A/Solids} \bibinfo{volume}{95}, \bibinfo{pages}{104639}.
\bibitem[{Abueidda et~al.(2021)Abueidda, Lu and Koric}]{abueidda2021meshless}
\bibinfo{author}{Abueidda, D.W.}, \bibinfo{author}{Lu, Q.}, \bibinfo{author}{Koric, S.}, \bibinfo{year}{2021}.
\newblock \bibinfo{title}{Meshless physics-informed deep learning method for three-dimensional solid mechanics}.
\newblock \bibinfo{journal}{International Journal for Numerical Methods in Engineering} \bibinfo{volume}{122}, \bibinfo{pages}{7182--7201}.
\bibitem[{Aln{\ae}s et~al.(2015)Aln{\ae}s, Blechta, Hake, Johansson, Kehlet, Logg, Richardson, Ring, Rognes and Wells}]{alnaes2015fenics}
\bibinfo{author}{Aln{\ae}s, M.}, \bibinfo{author}{Blechta, J.}, \bibinfo{author}{Hake, J.}, \bibinfo{author}{Johansson, A.}, \bibinfo{author}{Kehlet, B.}, \bibinfo{author}{Logg, A.}, \bibinfo{author}{Richardson, C.}, \bibinfo{author}{Ring, J.}, \bibinfo{author}{Rognes, M.E.}, \bibinfo{author}{Wells, G.N.}, \bibinfo{year}{2015}.
\newblock \bibinfo{title}{The fenics project version 1.5}.
\newblock \bibinfo{journal}{Archive of numerical software} \bibinfo{volume}{3}.
\bibitem[{Atluri and Zhu(2000)}]{atluri2000new}
\bibinfo{author}{Atluri, S.}, \bibinfo{author}{Zhu, T.}, \bibinfo{year}{2000}.
\newblock \bibinfo{title}{New concepts in meshless methods}.
\newblock \bibinfo{journal}{International journal for numerical methods in engineering} \bibinfo{volume}{47}, \bibinfo{pages}{537--556}.
\bibitem[{Atluri and Zhu(1998)}]{atluri1998new}
\bibinfo{author}{Atluri, S.N.}, \bibinfo{author}{Zhu, T.}, \bibinfo{year}{1998}.
\newblock \bibinfo{title}{A new meshless local petrov-galerkin (mlpg) approach in computational mechanics}.
\newblock \bibinfo{journal}{Computational mechanics} \bibinfo{volume}{22}, \bibinfo{pages}{117--127}.
\bibitem[{Babaee et~al.(2013)Babaee, Shim, Weaver, Chen, Patel and Bertoldi}]{babaee20133d}
\bibinfo{author}{Babaee, S.}, \bibinfo{author}{Shim, J.}, \bibinfo{author}{Weaver, J.C.}, \bibinfo{author}{Chen, E.R.}, \bibinfo{author}{Patel, N.}, \bibinfo{author}{Bertoldi, K.}, \bibinfo{year}{2013}.
\newblock \bibinfo{title}{3d soft metamaterials with negative poisson’s ratio}.
\newblock \bibinfo{journal}{Adv. Mater} \bibinfo{volume}{25}, \bibinfo{pages}{5044--5049}.
\bibitem[{Baek and Chen(2024)}]{baek2024neural}
\bibinfo{author}{Baek, J.}, \bibinfo{author}{Chen, J.S.}, \bibinfo{year}{2024}.
\newblock \bibinfo{title}{A neural network-based enrichment of reproducing kernel approximation for modeling brittle fracture}.
\newblock \bibinfo{journal}{Computer Methods in Applied Mechanics and Engineering} \bibinfo{volume}{419}, \bibinfo{pages}{116590}.
\bibitem[{Baek et~al.(2022)Baek, Chen and Susuki}]{baek2022neural}
\bibinfo{author}{Baek, J.}, \bibinfo{author}{Chen, J.S.}, \bibinfo{author}{Susuki, K.}, \bibinfo{year}{2022}.
\newblock \bibinfo{title}{A neural network-enhanced reproducing kernel particle method for modeling strain localization}.
\newblock \bibinfo{journal}{International Journal for Numerical Methods in Engineering} \bibinfo{volume}{123}, \bibinfo{pages}{4422--4454}.
\bibitem[{Baydin et~al.(2018)Baydin, Pearlmutter, Radul and Siskind}]{baydin2018automatic}
\bibinfo{author}{Baydin, A.G.}, \bibinfo{author}{Pearlmutter, B.A.}, \bibinfo{author}{Radul, A.A.}, \bibinfo{author}{Siskind, J.M.}, \bibinfo{year}{2018}.
\newblock \bibinfo{title}{Automatic differentiation in machine learning: a survey}.
\newblock \bibinfo{journal}{Journal of Marchine Learning Research} \bibinfo{volume}{18}, \bibinfo{pages}{1--43}.
\bibitem[{Beatty(1987)}]{beatty1987topics}
\bibinfo{author}{Beatty, M.F.}, \bibinfo{year}{1987}.
\newblock \bibinfo{title}{{Topics in Finite Elasticity: Hyperelasticity of Rubber, Elastomers, and Biological Tissues—With Examples}}.
\newblock \bibinfo{journal}{Applied Mechanics Reviews} \bibinfo{volume}{40}, \bibinfo{pages}{1699--1734}.
\bibitem[{Belytschko et~al.(2014)Belytschko, Liu, Moran and Elkhodary}]{belytschko2014nonlinear}
\bibinfo{author}{Belytschko, T.}, \bibinfo{author}{Liu, W.K.}, \bibinfo{author}{Moran, B.}, \bibinfo{author}{Elkhodary, K.}, \bibinfo{year}{2014}.
\newblock \bibinfo{title}{Nonlinear finite elements for continua and structures}.
\newblock \bibinfo{publisher}{John wiley \& sons}.
\bibitem[{Belytschko et~al.(1994)Belytschko, Lu and Gu}]{belytschko1994element}
\bibinfo{author}{Belytschko, T.}, \bibinfo{author}{Lu, Y.Y.}, \bibinfo{author}{Gu, L.}, \bibinfo{year}{1994}.
\newblock \bibinfo{title}{Element-free galerkin methods}.
\newblock \bibinfo{journal}{International journal for numerical methods in engineering} \bibinfo{volume}{37}, \bibinfo{pages}{229--256}.
\bibitem[{Berg and Nystr{\"o}m(2018)}]{berg2018unified}
\bibinfo{author}{Berg, J.}, \bibinfo{author}{Nystr{\"o}m, K.}, \bibinfo{year}{2018}.
\newblock \bibinfo{title}{A unified deep artificial neural network approach to partial differential equations in complex geometries}.
\newblock \bibinfo{journal}{Neurocomputing} \bibinfo{volume}{317}, \bibinfo{pages}{28--41}.
\bibitem[{Bezgin et~al.(2023)Bezgin, Buhendwa and Adams}]{bezgin2023jax}
\bibinfo{author}{Bezgin, D.A.}, \bibinfo{author}{Buhendwa, A.B.}, \bibinfo{author}{Adams, N.A.}, \bibinfo{year}{2023}.
\newblock \bibinfo{title}{Jax-fluids: A fully-differentiable high-order computational fluid dynamics solver for compressible two-phase flows}.
\newblock \bibinfo{journal}{Computer Physics Communications} \bibinfo{volume}{282}, \bibinfo{pages}{108527}.
\bibitem[{Blum and Li(1991)}]{blum1991approximation}
\bibinfo{author}{Blum, E.K.}, \bibinfo{author}{Li, L.K.}, \bibinfo{year}{1991}.
\newblock \bibinfo{title}{Approximation theory and feedforward networks}.
\newblock \bibinfo{journal}{Neural networks} \bibinfo{volume}{4}, \bibinfo{pages}{511--515}.
\bibitem[{Bradbury et~al.(2018)Bradbury, Frostig, Hawkins, Johnson, Leary, Maclaurin, Necula, Paszke, Vander{P}las, Wanderman-{M}ilne and Zhang}]{jax2018github}
\bibinfo{author}{Bradbury, J.}, \bibinfo{author}{Frostig, R.}, \bibinfo{author}{Hawkins, P.}, \bibinfo{author}{Johnson, M.J.}, \bibinfo{author}{Leary, C.}, \bibinfo{author}{Maclaurin, D.}, \bibinfo{author}{Necula, G.}, \bibinfo{author}{Paszke, A.}, \bibinfo{author}{Vander{P}las, J.}, \bibinfo{author}{Wanderman-{M}ilne, S.}, \bibinfo{author}{Zhang, Q.}, \bibinfo{year}{2018}.
\newblock \bibinfo{title}{{JAX}: composable transformations of {P}ython+{N}um{P}y programs}.
\newblock \URLprefix \url{http://github.com/google/jax}.
\bibitem[{Chagnon et~al.(2015)Chagnon, Rebouah and Favier}]{chagnon2015hyperelastic}
\bibinfo{author}{Chagnon, G.}, \bibinfo{author}{Rebouah, M.}, \bibinfo{author}{Favier, D.}, \bibinfo{year}{2015}.
\newblock \bibinfo{title}{Hyperelastic energy densities for soft biological tissues: a review}.
\newblock \bibinfo{journal}{Journal of Elasticity} \bibinfo{volume}{120}, \bibinfo{pages}{129--160}.
\bibitem[{Chen et~al.(2017)Chen, Hillman and Chi}]{chen2017meshfree}
\bibinfo{author}{Chen, J.S.}, \bibinfo{author}{Hillman, M.}, \bibinfo{author}{Chi, S.W.}, \bibinfo{year}{2017}.
\newblock \bibinfo{title}{Meshfree methods: progress made after 20 years}.
\newblock \bibinfo{journal}{Journal of Engineering Mechanics} \bibinfo{volume}{143}, \bibinfo{pages}{04017001}.
\bibitem[{Chen et~al.(1996)Chen, Pan, Wu and Liu}]{chen1996reproducing}
\bibinfo{author}{Chen, J.S.}, \bibinfo{author}{Pan, C.}, \bibinfo{author}{Wu, C.T.}, \bibinfo{author}{Liu, W.K.}, \bibinfo{year}{1996}.
\newblock \bibinfo{title}{Reproducing kernel particle methods for large deformation analysis of non-linear structures}.
\newblock \bibinfo{journal}{Computer methods in applied mechanics and engineering} \bibinfo{volume}{139}, \bibinfo{pages}{195--227}.
\bibitem[{Chen and Wang(2000)}]{chen2000new}
\bibinfo{author}{Chen, J.S.}, \bibinfo{author}{Wang, H.P.}, \bibinfo{year}{2000}.
\newblock \bibinfo{title}{New boundary condition treatments in meshfree computation of contact problems}.
\newblock \bibinfo{journal}{Computer methods in applied mechanics and engineering} \bibinfo{volume}{187}, \bibinfo{pages}{441--468}.
\bibitem[{Dong et~al.(2023)Dong, Liu, Li and Qiao}]{dong2023deepfem}
\bibinfo{author}{Dong, Y.}, \bibinfo{author}{Liu, T.}, \bibinfo{author}{Li, Z.}, \bibinfo{author}{Qiao, P.}, \bibinfo{year}{2023}.
\newblock \bibinfo{title}{Deepfem: A novel element-based deep learning approach for solving nonlinear partial differential equations in computational solid mechanics}.
\newblock \bibinfo{journal}{Journal of Engineering Mechanics} \bibinfo{volume}{149}, \bibinfo{pages}{04022102}.
\bibitem[{Du and He(2024)}]{du2024neural}
\bibinfo{author}{Du, H.}, \bibinfo{author}{He, Q.}, \bibinfo{year}{2024}.
\newblock \bibinfo{title}{Neural-integrated meshfree (nim) method: A differentiable programming-based hybrid solver for computational mechanics}.
\newblock \bibinfo{journal}{Computer Methods in Applied Mechanics and Engineering} \bibinfo{volume}{427}, \bibinfo{pages}{117024}.
\bibitem[{Du et~al.(2023)Du, Zhao, Cheng, Yan and He}]{du2023modeling}
\bibinfo{author}{Du, H.}, \bibinfo{author}{Zhao, Z.}, \bibinfo{author}{Cheng, H.}, \bibinfo{author}{Yan, J.}, \bibinfo{author}{He, Q.}, \bibinfo{year}{2023}.
\newblock \bibinfo{title}{Modeling density-driven flow in porous media by physics-informed neural networks for co2 sequestration}.
\newblock \bibinfo{journal}{Computers and Geotechnics} \bibinfo{volume}{159}, \bibinfo{pages}{105433}.
\bibitem[{Duarte and Oden(1996)}]{duarte1996hp}
\bibinfo{author}{Duarte, C.A.}, \bibinfo{author}{Oden, J.T.}, \bibinfo{year}{1996}.
\newblock \bibinfo{title}{An hp adaptive method using clouds}.
\newblock \bibinfo{journal}{Computer methods in applied mechanics and engineering} \bibinfo{volume}{139}, \bibinfo{pages}{237--262}.
\bibitem[{Fang(2021)}]{fang2021high}
\bibinfo{author}{Fang, Z.}, \bibinfo{year}{2021}.
\newblock \bibinfo{title}{A high-efficient hybrid physics-informed neural networks based on convolutional neural network}.
\newblock \bibinfo{journal}{IEEE Transactions on Neural Networks and Learning Systems} \bibinfo{volume}{33}, \bibinfo{pages}{5514--5526}.
\bibitem[{Fuhg and Bouklas(2022)}]{fuhg2022mixed}
\bibinfo{author}{Fuhg, J.N.}, \bibinfo{author}{Bouklas, N.}, \bibinfo{year}{2022}.
\newblock \bibinfo{title}{The mixed deep energy method for resolving concentration features in finite strain hyperelasticity}.
\newblock \bibinfo{journal}{Journal of Computational Physics} \bibinfo{volume}{451}, \bibinfo{pages}{110839}.
\bibitem[{Fuhg et~al.(2024)Fuhg, Padmanabha, Bouklas, Bahmani, Sun, Vlassis, Flaschel, Carrara and De~Lorenzis}]{fuhg2024review}
\bibinfo{author}{Fuhg, J.N.}, \bibinfo{author}{Padmanabha, G.A.}, \bibinfo{author}{Bouklas, N.}, \bibinfo{author}{Bahmani, B.}, \bibinfo{author}{Sun, W.}, \bibinfo{author}{Vlassis, N.N.}, \bibinfo{author}{Flaschel, M.}, \bibinfo{author}{Carrara, P.}, \bibinfo{author}{De~Lorenzis, L.}, \bibinfo{year}{2024}.
\newblock \bibinfo{title}{A review on data-driven constitutive laws for solids}.
\newblock \bibinfo{journal}{arXiv preprint arXiv:2405.03658} .
\bibitem[{Gao et~al.(2022)Gao, Zahr and Wang}]{gao2022physics}
\bibinfo{author}{Gao, H.}, \bibinfo{author}{Zahr, M.J.}, \bibinfo{author}{Wang, J.X.}, \bibinfo{year}{2022}.
\newblock \bibinfo{title}{Physics-informed graph neural galerkin networks: A unified framework for solving pde-governed forward and inverse problems}.
\newblock \bibinfo{journal}{Computer Methods in Applied Mechanics and Engineering} \bibinfo{volume}{390}, \bibinfo{pages}{114502}.
\bibitem[{Gasick and Qian(2023)}]{gasick2023isogeometric}
\bibinfo{author}{Gasick, J.}, \bibinfo{author}{Qian, X.}, \bibinfo{year}{2023}.
\newblock \bibinfo{title}{Isogeometric neural networks: A new deep learning approach for solving parameterized partial differential equations}.
\newblock \bibinfo{journal}{Computer Methods in Applied Mechanics and Engineering} \bibinfo{volume}{405}, \bibinfo{pages}{115839}.
\bibitem[{Haghighat et~al.(2021)Haghighat, Raissi, Moure, Gomez and Juanes}]{haghighat2021physics}
\bibinfo{author}{Haghighat, E.}, \bibinfo{author}{Raissi, M.}, \bibinfo{author}{Moure, A.}, \bibinfo{author}{Gomez, H.}, \bibinfo{author}{Juanes, R.}, \bibinfo{year}{2021}.
\newblock \bibinfo{title}{A physics-informed deep learning framework for inversion and surrogate modeling in solid mechanics}.
\newblock \bibinfo{journal}{Computer Methods in Applied Mechanics and Engineering} \bibinfo{volume}{379}, \bibinfo{pages}{113741}.
\bibitem[{He et~al.(2023)He, Abueidda, Al-Rub, Koric and Jasiuk}]{he2023deep}
\bibinfo{author}{He, J.}, \bibinfo{author}{Abueidda, D.}, \bibinfo{author}{Al-Rub, R.A.}, \bibinfo{author}{Koric, S.}, \bibinfo{author}{Jasiuk, I.}, \bibinfo{year}{2023}.
\newblock \bibinfo{title}{A deep learning energy-based method for classical elastoplasticity}.
\newblock \bibinfo{journal}{International Journal of Plasticity} \bibinfo{volume}{162}, \bibinfo{pages}{103531}.
\bibitem[{He et~al.(2024)He, Zhao, Cheung, Zhang and Ren}]{HE2024106443}
\bibinfo{author}{He, L.}, \bibinfo{author}{Zhao, M.}, \bibinfo{author}{Cheung, J.P.Y.}, \bibinfo{author}{Zhang, T.}, \bibinfo{author}{Ren, X.}, \bibinfo{year}{2024}.
\newblock \bibinfo{title}{Gaussian random field-based characterization and reconstruction of cancellous bone microstructure considering the constraint of correlation structure}.
\newblock \bibinfo{journal}{Journal of the Mechanical Behavior of Biomedical Materials} \bibinfo{volume}{152}, \bibinfo{pages}{106443}.
\newblock \URLprefix \url{https://www.sciencedirect.com/science/article/pii/S1751616124000754}, \DOIprefix\doi{https://doi.org/10.1016/j.jmbbm.2024.106443}.
\bibitem[{He et~al.(2020)He, Barajas-Solano, Tartakovsky and Tartakovsky}]{he2020physics}
\bibinfo{author}{He, Q.}, \bibinfo{author}{Barajas-Solano, D.}, \bibinfo{author}{Tartakovsky, G.}, \bibinfo{author}{Tartakovsky, A.M.}, \bibinfo{year}{2020}.
\newblock \bibinfo{title}{Physics-informed neural networks for multiphysics data assimilation with application to subsurface transport}.
\newblock \bibinfo{journal}{Advances in Water Resources} \bibinfo{volume}{141}, \bibinfo{pages}{103610}.
\bibitem[{He and Chen(2020)}]{he2020physics_lcdd}
\bibinfo{author}{He, Q.}, \bibinfo{author}{Chen, J.S.}, \bibinfo{year}{2020}.
\newblock \bibinfo{title}{A physics-constrained data-driven approach based on locally convex reconstruction for noisy database}.
\newblock \bibinfo{journal}{Computer Methods in Applied Mechanics and Engineering} \bibinfo{volume}{363}, \bibinfo{pages}{112791}.
\bibitem[{He et~al.(2021)He, Laurence, Lee and Chen}]{he2021manifold}
\bibinfo{author}{He, Q.}, \bibinfo{author}{Laurence, D.W.}, \bibinfo{author}{Lee, C.H.}, \bibinfo{author}{Chen, J.S.}, \bibinfo{year}{2021}.
\newblock \bibinfo{title}{Manifold learning based data-driven modeling for soft biological tissues}.
\newblock \bibinfo{journal}{Journal of biomechanics} \bibinfo{volume}{117}, \bibinfo{pages}{110124}.
\bibitem[{He and Tartakovsky(2021)}]{he2021physics}
\bibinfo{author}{He, Q.}, \bibinfo{author}{Tartakovsky, A.M.}, \bibinfo{year}{2021}.
\newblock \bibinfo{title}{Physics-informed neural network method for forward and backward advection-dispersion equations}.
\newblock \bibinfo{journal}{Water Resources Research} \bibinfo{volume}{57}, \bibinfo{pages}{e2020WR029479}.
\bibitem[{Holzapfel(2002)}]{holzapfel2002nonlinear}
\bibinfo{author}{Holzapfel, G.A.}, \bibinfo{year}{2002}.
\newblock \bibinfo{title}{Nonlinear solid mechanics: a continuum approach for engineering science}.
\bibitem[{Hornik(1991)}]{hornik1991approximation}
\bibinfo{author}{Hornik, K.}, \bibinfo{year}{1991}.
\newblock \bibinfo{title}{Approximation capabilities of multilayer feedforward networks}.
\newblock \bibinfo{journal}{Neural networks} \bibinfo{volume}{4}, \bibinfo{pages}{251--257}.
\bibitem[{Innes et~al.(2019)Innes, Edelman, Fischer, Rackauckas, Saba, Shah and Tebbutt}]{innes2019differentiable}
\bibinfo{author}{Innes, M.}, \bibinfo{author}{Edelman, A.}, \bibinfo{author}{Fischer, K.}, \bibinfo{author}{Rackauckas, C.}, \bibinfo{author}{Saba, E.}, \bibinfo{author}{Shah, V.B.}, \bibinfo{author}{Tebbutt, W.}, \bibinfo{year}{2019}.
\newblock \bibinfo{title}{A differentiable programming system to bridge machine learning and scientific computing}.
\newblock \bibinfo{journal}{arXiv preprint arXiv:1907.07587} .
\bibitem[{Karniadakis et~al.(2021)Karniadakis, Kevrekidis, Lu, Perdikaris, Wang and Yang}]{karniadakis2021physics}
\bibinfo{author}{Karniadakis, G.E.}, \bibinfo{author}{Kevrekidis, I.G.}, \bibinfo{author}{Lu, L.}, \bibinfo{author}{Perdikaris, P.}, \bibinfo{author}{Wang, S.}, \bibinfo{author}{Yang, L.}, \bibinfo{year}{2021}.
\newblock \bibinfo{title}{Physics-informed machine learning}.
\newblock \bibinfo{journal}{Nature Reviews Physics} \bibinfo{volume}{3}, \bibinfo{pages}{422--440}.
\bibitem[{Kashinath et~al.(2021)Kashinath, Mustafa, Albert, Wu, Jiang, Esmaeilzadeh, Azizzadenesheli, Wang, Chattopadhyay, Singh et~al.}]{kashinath2021physics}
\bibinfo{author}{Kashinath, K.}, \bibinfo{author}{Mustafa, M.}, \bibinfo{author}{Albert, A.}, \bibinfo{author}{Wu, J.}, \bibinfo{author}{Jiang, C.}, \bibinfo{author}{Esmaeilzadeh, S.}, \bibinfo{author}{Azizzadenesheli, K.}, \bibinfo{author}{Wang, R.}, \bibinfo{author}{Chattopadhyay, A.}, \bibinfo{author}{Singh, A.}, et~al., \bibinfo{year}{2021}.
\newblock \bibinfo{title}{Physics-informed machine learning: case studies for weather and climate modelling}.
\newblock \bibinfo{journal}{Philosophical Transactions of the Royal Society A} \bibinfo{volume}{379}, \bibinfo{pages}{20200093}.
\bibitem[{Khara et~al.(2024)Khara, Balu, Joshi, Sarkar, Hegde, Krishnamurthy and Ganapathysubramanian}]{khara2024neufenet}
\bibinfo{author}{Khara, B.}, \bibinfo{author}{Balu, A.}, \bibinfo{author}{Joshi, A.}, \bibinfo{author}{Sarkar, S.}, \bibinfo{author}{Hegde, C.}, \bibinfo{author}{Krishnamurthy, A.}, \bibinfo{author}{Ganapathysubramanian, B.}, \bibinfo{year}{2024}.
\newblock \bibinfo{title}{Neufenet: Neural finite element solutions with theoretical bounds for parametric pdes}.
\newblock \bibinfo{journal}{Engineering with Computers} , \bibinfo{pages}{1--23}.
\bibitem[{Kharazmi et~al.(2021)Kharazmi, Zhang and Karniadakis}]{kharazmi2021hp}
\bibinfo{author}{Kharazmi, E.}, \bibinfo{author}{Zhang, Z.}, \bibinfo{author}{Karniadakis, G.E.}, \bibinfo{year}{2021}.
\newblock \bibinfo{title}{hp-vpinns: Variational physics-informed neural networks with domain decomposition}.
\newblock \bibinfo{journal}{Computer Methods in Applied Mechanics and Engineering} \bibinfo{volume}{374}, \bibinfo{pages}{113547}.
\bibitem[{Khodayi-Mehr and Zavlanos(2020)}]{khodayi2020varnet}
\bibinfo{author}{Khodayi-Mehr, R.}, \bibinfo{author}{Zavlanos, M.}, \bibinfo{year}{2020}.
\newblock \bibinfo{title}{Varnet: Variational neural networks for the solution of partial differential equations}, in: \bibinfo{booktitle}{Learning for Dynamics and Control}, \bibinfo{organization}{PMLR}. pp. \bibinfo{pages}{298--307}.
\bibitem[{Kirchdoerfer and Ortiz(2016)}]{kirchdoerfer2016data}
\bibinfo{author}{Kirchdoerfer, T.}, \bibinfo{author}{Ortiz, M.}, \bibinfo{year}{2016}.
\newblock \bibinfo{title}{Data-driven computational mechanics}.
\newblock \bibinfo{journal}{Computer Methods in Applied Mechanics and Engineering} \bibinfo{volume}{304}, \bibinfo{pages}{81--101}.
\bibitem[{Krishnapriyan et~al.(2021)Krishnapriyan, Gholami, Zhe, Kirby and Mahoney}]{krishnapriyan2021characterizing}
\bibinfo{author}{Krishnapriyan, A.}, \bibinfo{author}{Gholami, A.}, \bibinfo{author}{Zhe, S.}, \bibinfo{author}{Kirby, R.}, \bibinfo{author}{Mahoney, M.W.}, \bibinfo{year}{2021}.
\newblock \bibinfo{title}{Characterizing possible failure modes in physics-informed neural networks}.
\newblock \bibinfo{journal}{Advances in Neural Information Processing Systems} \bibinfo{volume}{34}, \bibinfo{pages}{26548--26560}.
\bibitem[{Lagaris et~al.(1998)Lagaris, Likas and Fotiadis}]{lagaris1998artificial}
\bibinfo{author}{Lagaris, I.E.}, \bibinfo{author}{Likas, A.}, \bibinfo{author}{Fotiadis, D.I.}, \bibinfo{year}{1998}.
\newblock \bibinfo{title}{Artificial neural networks for solving ordinary and partial differential equations}.
\newblock \bibinfo{journal}{IEEE transactions on neural networks} \bibinfo{volume}{9}, \bibinfo{pages}{987--1000}.
\bibitem[{Lee et~al.(2021)Lee, Trask, Patel, Gulian and Cyr}]{lee2021partition}
\bibinfo{author}{Lee, K.}, \bibinfo{author}{Trask, N.A.}, \bibinfo{author}{Patel, R.G.}, \bibinfo{author}{Gulian, M.A.}, \bibinfo{author}{Cyr, E.C.}, \bibinfo{year}{2021}.
\newblock \bibinfo{title}{Partition of unity networks: deep hp-approximation}.
\newblock \bibinfo{journal}{arXiv preprint arXiv:2101.11256} .
\bibitem[{Linka et~al.(2021)Linka, Hillg{\"a}rtner, Abdolazizi, Aydin, Itskov and Cyron}]{linka2021constitutive}
\bibinfo{author}{Linka, K.}, \bibinfo{author}{Hillg{\"a}rtner, M.}, \bibinfo{author}{Abdolazizi, K.P.}, \bibinfo{author}{Aydin, R.C.}, \bibinfo{author}{Itskov, M.}, \bibinfo{author}{Cyron, C.J.}, \bibinfo{year}{2021}.
\newblock \bibinfo{title}{Constitutive artificial neural networks: A fast and general approach to predictive data-driven constitutive modeling by deep learning}.
\newblock \bibinfo{journal}{Journal of Computational Physics} \bibinfo{volume}{429}, \bibinfo{pages}{110010}.
\bibitem[{Liu et~al.(1995)Liu, Jun and Zhang}]{liu1995reproducing}
\bibinfo{author}{Liu, W.K.}, \bibinfo{author}{Jun, S.}, \bibinfo{author}{Zhang, Y.F.}, \bibinfo{year}{1995}.
\newblock \bibinfo{title}{Reproducing kernel particle methods}.
\newblock \bibinfo{journal}{International journal for numerical methods in fluids} \bibinfo{volume}{20}, \bibinfo{pages}{1081--1106}.
\bibitem[{Liu et~al.(2019)Liu, Wu and Koishi}]{liu2019deep}
\bibinfo{author}{Liu, Z.}, \bibinfo{author}{Wu, C.}, \bibinfo{author}{Koishi, M.}, \bibinfo{year}{2019}.
\newblock \bibinfo{title}{A deep material network for multiscale topology learning and accelerated nonlinear modeling of heterogeneous materials}.
\newblock \bibinfo{journal}{Computer Methods in Applied Mechanics and Engineering} \bibinfo{volume}{345}, \bibinfo{pages}{1138--1168}.
\bibitem[{Lu et~al.(2021)Lu, Jin, Pang, Zhang and Karniadakis}]{lu2021learning}
\bibinfo{author}{Lu, L.}, \bibinfo{author}{Jin, P.}, \bibinfo{author}{Pang, G.}, \bibinfo{author}{Zhang, Z.}, \bibinfo{author}{Karniadakis, G.E.}, \bibinfo{year}{2021}.
\newblock \bibinfo{title}{Learning nonlinear operators via deeponet based on the universal approximation theorem of operators}.
\newblock \bibinfo{journal}{Nature machine intelligence} \bibinfo{volume}{3}, \bibinfo{pages}{218--229}.
\bibitem[{Lu et~al.(2023)Lu, Li, Zhang, Park, Mojumder, Knapik, Sang, Tang, Apley, Wagner et~al.}]{lu2023convolution}
\bibinfo{author}{Lu, Y.}, \bibinfo{author}{Li, H.}, \bibinfo{author}{Zhang, L.}, \bibinfo{author}{Park, C.}, \bibinfo{author}{Mojumder, S.}, \bibinfo{author}{Knapik, S.}, \bibinfo{author}{Sang, Z.}, \bibinfo{author}{Tang, S.}, \bibinfo{author}{Apley, D.W.}, \bibinfo{author}{Wagner, G.J.}, et~al., \bibinfo{year}{2023}.
\newblock \bibinfo{title}{Convolution hierarchical deep-learning neural networks (c-hidenn): finite elements, isogeometric analysis, tensor decomposition, and beyond}.
\newblock \bibinfo{journal}{Computational Mechanics} \bibinfo{volume}{72}, \bibinfo{pages}{333--362}.
\bibitem[{Masi and Stefanou(2022)}]{masi2022multiscale}
\bibinfo{author}{Masi, F.}, \bibinfo{author}{Stefanou, I.}, \bibinfo{year}{2022}.
\newblock \bibinfo{title}{Multiscale modeling of inelastic materials with thermodynamics-based artificial neural networks (tann)}.
\newblock \bibinfo{journal}{Computer Methods in Applied Mechanics and Engineering} \bibinfo{volume}{398}, \bibinfo{pages}{115190}.
\bibitem[{Masi et~al.(2021)Masi, Stefanou, Vannucci and Maffi-Berthier}]{masi2021thermodynamics}
\bibinfo{author}{Masi, F.}, \bibinfo{author}{Stefanou, I.}, \bibinfo{author}{Vannucci, P.}, \bibinfo{author}{Maffi-Berthier, V.}, \bibinfo{year}{2021}.
\newblock \bibinfo{title}{Thermodynamics-based artificial neural networks for constitutive modeling}.
\newblock \bibinfo{journal}{Journal of the Mechanics and Physics of Solids} \bibinfo{volume}{147}, \bibinfo{pages}{104277}.
\bibitem[{Melenk and Babu{\v{s}}ka(1996)}]{melenk1996partition}
\bibinfo{author}{Melenk, J.M.}, \bibinfo{author}{Babu{\v{s}}ka, I.}, \bibinfo{year}{1996}.
\newblock \bibinfo{title}{The partition of unity finite element method: basic theory and applications}.
\newblock \bibinfo{journal}{Computer methods in applied mechanics and engineering} \bibinfo{volume}{139}, \bibinfo{pages}{289--314}.
\bibitem[{Nayroles et~al.(1992)Nayroles, Touzot and Villon}]{nayroles1992generalizing}
\bibinfo{author}{Nayroles, B.}, \bibinfo{author}{Touzot, G.}, \bibinfo{author}{Villon, P.}, \bibinfo{year}{1992}.
\newblock \bibinfo{title}{Generalizing the finite element method: diffuse approximation and diffuse elements}.
\newblock \bibinfo{journal}{Computational mechanics} \bibinfo{volume}{10}, \bibinfo{pages}{307--318}.
\bibitem[{Nguyen-Thanh et~al.(2020)Nguyen-Thanh, Zhuang and Rabczuk}]{nguyen2020deep}
\bibinfo{author}{Nguyen-Thanh, V.M.}, \bibinfo{author}{Zhuang, X.}, \bibinfo{author}{Rabczuk, T.}, \bibinfo{year}{2020}.
\newblock \bibinfo{title}{A deep energy method for finite deformation hyperelasticity}.
\newblock \bibinfo{journal}{European Journal of Mechanics-A/Solids} \bibinfo{volume}{80}, \bibinfo{pages}{103874}.
\bibitem[{Niu et~al.(2023)Niu, Zhang, Bazilevs and Srivastava}]{niu2023modeling}
\bibinfo{author}{Niu, S.}, \bibinfo{author}{Zhang, E.}, \bibinfo{author}{Bazilevs, Y.}, \bibinfo{author}{Srivastava, V.}, \bibinfo{year}{2023}.
\newblock \bibinfo{title}{Modeling finite-strain plasticity using physics-informed neural network and assessment of the network performance}.
\newblock \bibinfo{journal}{Journal of the Mechanics and Physics of Solids} \bibinfo{volume}{172}, \bibinfo{pages}{105177}.
\bibitem[{Park et~al.(2023)Park, Lu, Saha, Xue, Guo, Mojumder, Apley, Wagner and Liu}]{park2023convolution}
\bibinfo{author}{Park, C.}, \bibinfo{author}{Lu, Y.}, \bibinfo{author}{Saha, S.}, \bibinfo{author}{Xue, T.}, \bibinfo{author}{Guo, J.}, \bibinfo{author}{Mojumder, S.}, \bibinfo{author}{Apley, D.W.}, \bibinfo{author}{Wagner, G.J.}, \bibinfo{author}{Liu, W.K.}, \bibinfo{year}{2023}.
\newblock \bibinfo{title}{Convolution hierarchical deep-learning neural network (c-hidenn) with graphics processing unit (gpu) acceleration}.
\newblock \bibinfo{journal}{Computational Mechanics} \bibinfo{volume}{72}, \bibinfo{pages}{383--409}.
\bibitem[{Paszke et~al.(2019)Paszke, Gross, Massa, Lerer, Bradbury, Chanan, Killeen, Lin, Gimelshein, Antiga et~al.}]{paszke2019pytorch}
\bibinfo{author}{Paszke, A.}, \bibinfo{author}{Gross, S.}, \bibinfo{author}{Massa, F.}, \bibinfo{author}{Lerer, A.}, \bibinfo{author}{Bradbury, J.}, \bibinfo{author}{Chanan, G.}, \bibinfo{author}{Killeen, T.}, \bibinfo{author}{Lin, Z.}, \bibinfo{author}{Gimelshein, N.}, \bibinfo{author}{Antiga, L.}, et~al., \bibinfo{year}{2019}.
\newblock \bibinfo{title}{Pytorch: An imperative style, high-performance deep learning library}.
\newblock \bibinfo{journal}{Advances in neural information processing systems} \bibinfo{volume}{32}.
\bibitem[{Raissi et~al.(2019)Raissi, Perdikaris and Karniadakis}]{raissi2019physics}
\bibinfo{author}{Raissi, M.}, \bibinfo{author}{Perdikaris, P.}, \bibinfo{author}{Karniadakis, G.E.}, \bibinfo{year}{2019}.
\newblock \bibinfo{title}{Physics-informed neural networks: A deep learning framework for solving forward and inverse problems involving nonlinear partial differential equations}.
\newblock \bibinfo{journal}{Journal of Computational physics} \bibinfo{volume}{378}, \bibinfo{pages}{686--707}.
\bibitem[{Rao et~al.(2021)Rao, Sun and Liu}]{rao2021physics}
\bibinfo{author}{Rao, C.}, \bibinfo{author}{Sun, H.}, \bibinfo{author}{Liu, Y.}, \bibinfo{year}{2021}.
\newblock \bibinfo{title}{Physics-informed deep learning for computational elastodynamics without labeled data}.
\newblock \bibinfo{journal}{Journal of Engineering Mechanics} \bibinfo{volume}{147}, \bibinfo{pages}{04021043}.
\bibitem[{Rezaei et~al.(2022)Rezaei, Harandi, Moeineddin, Xu and Reese}]{rezaei2022mixed}
\bibinfo{author}{Rezaei, S.}, \bibinfo{author}{Harandi, A.}, \bibinfo{author}{Moeineddin, A.}, \bibinfo{author}{Xu, B.X.}, \bibinfo{author}{Reese, S.}, \bibinfo{year}{2022}.
\newblock \bibinfo{title}{A mixed formulation for physics-informed neural networks as a potential solver for engineering problems in heterogeneous domains: Comparison with finite element method}.
\newblock \bibinfo{journal}{Computer Methods in Applied Mechanics and Engineering} \bibinfo{volume}{401}, \bibinfo{pages}{115616}.
\bibitem[{Samaniego et~al.(2020)Samaniego, Anitescu, Goswami, Nguyen-Thanh, Guo, Hamdia, Zhuang and Rabczuk}]{samaniego2020energy}
\bibinfo{author}{Samaniego, E.}, \bibinfo{author}{Anitescu, C.}, \bibinfo{author}{Goswami, S.}, \bibinfo{author}{Nguyen-Thanh, V.M.}, \bibinfo{author}{Guo, H.}, \bibinfo{author}{Hamdia, K.}, \bibinfo{author}{Zhuang, X.}, \bibinfo{author}{Rabczuk, T.}, \bibinfo{year}{2020}.
\newblock \bibinfo{title}{An energy approach to the solution of partial differential equations in computational mechanics via machine learning: Concepts, implementation and applications}.
\newblock \bibinfo{journal}{Computer Methods in Applied Mechanics and Engineering} \bibinfo{volume}{362}, \bibinfo{pages}{112790}.
\bibitem[{Sukumar and Srivastava(2022)}]{SUKUMAR2022114333}
\bibinfo{author}{Sukumar, N.}, \bibinfo{author}{Srivastava, A.}, \bibinfo{year}{2022}.
\newblock \bibinfo{title}{Exact imposition of boundary conditions with distance functions in physics-informed deep neural networks}.
\newblock \bibinfo{journal}{Computer Methods in Applied Mechanics and Engineering} \bibinfo{volume}{389}, \bibinfo{pages}{114333}.
\newblock \URLprefix \url{https://www.sciencedirect.com/science/article/pii/S0045782521006186}, \DOIprefix\doi{https://doi.org/10.1016/j.cma.2021.114333}.
\bibitem[{Vlassis et~al.(2020)Vlassis, Ma and Sun}]{vlassis2020geometric}
\bibinfo{author}{Vlassis, N.N.}, \bibinfo{author}{Ma, R.}, \bibinfo{author}{Sun, W.}, \bibinfo{year}{2020}.
\newblock \bibinfo{title}{Geometric deep learning for computational mechanics part i: Anisotropic hyperelasticity}.
\newblock \bibinfo{journal}{Computer Methods in Applied Mechanics and Engineering} \bibinfo{volume}{371}, \bibinfo{pages}{113299}.
\bibitem[{Wriggers(2008)}]{wriggers2008nonlinear}
\bibinfo{author}{Wriggers, P.}, \bibinfo{year}{2008}.
\newblock \bibinfo{title}{Nonlinear finite element methods}.
\newblock \bibinfo{publisher}{Springer Science \& Business Media}.
\bibitem[{Wu et~al.(2018)Wu, Xiao and Paterson}]{wu2018physics}
\bibinfo{author}{Wu, J.L.}, \bibinfo{author}{Xiao, H.}, \bibinfo{author}{Paterson, E.}, \bibinfo{year}{2018}.
\newblock \bibinfo{title}{Physics-informed machine learning approach for augmenting turbulence models: A comprehensive framework}.
\newblock \bibinfo{journal}{Physical Review Fluids} \bibinfo{volume}{3}, \bibinfo{pages}{074602}.
\bibitem[{Wu et~al.(2024)Wu, Daneker, Turner, Jolley and Lu}]{Wu2024-bg}
\bibinfo{author}{Wu, W.}, \bibinfo{author}{Daneker, M.}, \bibinfo{author}{Turner, K.T.}, \bibinfo{author}{Jolley, M.A.}, \bibinfo{author}{Lu, L.}, \bibinfo{year}{2024}.
\newblock \bibinfo{title}{Identifying heterogeneous micromechanical properties of biological tissues via physics-informed neural networks}.
\newblock \bibinfo{journal}{arXiv preprint arXiv:2402.10741} .
\bibitem[{Xue et~al.(2023)Xue, Liao, Gan, Park, Xie, Liu and Cao}]{xue2023jax}
\bibinfo{author}{Xue, T.}, \bibinfo{author}{Liao, S.}, \bibinfo{author}{Gan, Z.}, \bibinfo{author}{Park, C.}, \bibinfo{author}{Xie, X.}, \bibinfo{author}{Liu, W.K.}, \bibinfo{author}{Cao, J.}, \bibinfo{year}{2023}.
\newblock \bibinfo{title}{Jax-fem: A differentiable gpu-accelerated 3d finite element solver for automatic inverse design and mechanistic data science}.
\newblock \bibinfo{journal}{Computer Physics Communications} , \bibinfo{pages}{108802}.
\bibitem[{Yu et~al.(2018)}]{yu2018deep}
\bibinfo{author}{Yu, B.}, et~al., \bibinfo{year}{2018}.
\newblock \bibinfo{title}{The deep ritz method: a deep learning-based numerical algorithm for solving variational problems}.
\newblock \bibinfo{journal}{Communications in Mathematics and Statistics} \bibinfo{volume}{6}, \bibinfo{pages}{1--12}.
\bibitem[{Zhang et~al.(2022)Zhang, Dao, Karniadakis and Suresh}]{zhang2022analyses}
\bibinfo{author}{Zhang, E.}, \bibinfo{author}{Dao, M.}, \bibinfo{author}{Karniadakis, G.E.}, \bibinfo{author}{Suresh, S.}, \bibinfo{year}{2022}.
\newblock \bibinfo{title}{Analyses of internal structures and defects in materials using physics-informed neural networks}.
\newblock \bibinfo{journal}{Science advances} \bibinfo{volume}{8}, \bibinfo{pages}{eabk0644}.
\bibitem[{Zhang et~al.(2021)Zhang, Cheng, Li, Gao, Yu, Domel, Yang, Tang and Liu}]{zhang2021hierarchical}
\bibinfo{author}{Zhang, L.}, \bibinfo{author}{Cheng, L.}, \bibinfo{author}{Li, H.}, \bibinfo{author}{Gao, J.}, \bibinfo{author}{Yu, C.}, \bibinfo{author}{Domel, R.}, \bibinfo{author}{Yang, Y.}, \bibinfo{author}{Tang, S.}, \bibinfo{author}{Liu, W.K.}, \bibinfo{year}{2021}.
\newblock \bibinfo{title}{Hierarchical deep-learning neural networks: finite elements and beyond}.
\newblock \bibinfo{journal}{Computational Mechanics} \bibinfo{volume}{67}, \bibinfo{pages}{207--230}.
\bibitem[{Zobeiry and Humfeld(2021)}]{zobeiry2021physics}
\bibinfo{author}{Zobeiry, N.}, \bibinfo{author}{Humfeld, K.D.}, \bibinfo{year}{2021}.
\newblock \bibinfo{title}{A physics-informed machine learning approach for solving heat transfer equation in advanced manufacturing and engineering applications}.
\newblock \bibinfo{journal}{Engineering Applications of Artificial Intelligence} \bibinfo{volume}{101}, \bibinfo{pages}{104232}.

\end{thebibliography}

\end{document}